# Underwater Image Enhancement with Physical-based Denoising Diffusion Implicit Models

Nguyen Gia Bach [1], Chanh Minh Tran [2], Eiji Kamioka [1] and Phan Xuan Tan [1,*]

[1] Graduate School of Engineering and Science, Shibaura Institute of Technology, Japan
[2] College of Engineering, Shibaura Institute of Technology, Japan
Email: nb23505@shibaura-it.ac.jp (N.G.B.); tran.chanh.r4@sic.shibaura-it.ac.jp (C.M.T.); kamioka@shibaura-it.ac.jp (E.K.); tanpx@shibaura-it.ac.jp (P.X.T)
*Corresponding author

***Abstract*— Underwater vision is crucial for autonomous underwater vehicles (AUVs), and enhancing degraded underwater images in real-time on a resource-constrained AUV is a key challenge due to factors like light absorption and scattering, or the sufficient model computational complexity to resolve such factors. Traditional image enhancement techniques lack adaptability to varying underwater conditions, while learning-based methods, particularly those using convolutional neural networks (CNNs) and generative adversarial networks (GANs), offer more robust solutions but face limitations such as inadequate enhancement, unstable training, or mode collapse. Denoising diffusion probabilistic models (DDPMs) have emerged as a state-of-the-art approach in image-to-image tasks but require intensive computational complexity to achieve the desired underwater image enhancement (UIE) using the recent UW-DDPM solution. To address these challenges, this paper introduces UW-DiffPhys, a novel physical-based and diffusion-based UIE approach. UW-DiffPhys combines light-computation physical-based UIE network components with a denoising U-Net to replace the computationally intensive distribution transformation U-Net in the existing UW-DDPM framework, reducing complexity while maintaining performance. Additionally, the Denoising Diffusion Implicit Model (DDIM) is employed to accelerate the inference process through non-Markovian sampling. Experimental results demonstrate that UW-DiffPhys achieved a substantial reduction in computational complexity and inference time compared to UW-DDPM, with competitive performance in key metrics such as PSNR, SSIM, UCIQE, and an improvement in the overall underwater image quality UIQM metric. The implementation code can be found at the following repository: https://github.com/bachzz/UW-DiffPhys**

## I. Introduction

Underwater vision has become critical for autonomous underwater vehicles (AUVs) to execute tasks in the marine environment, with increasing transition from using sonar, laser, and infrared sensors to using visual sensor [1]. These underwater tasks can range from low-level vision-based localization and navigation of AUV [2], to high-level tasks such as marine biology and archaeology exploration [3], underwater target detection [4], underwater surveying and mapping [5], and so on. Due to underwater attenuation effects caused by light absorption and scattering, these applications require AUV to perform enhancement of the degraded underwater images in real-time, as the first step before executing any downstream tasks [6]. Therefore, how to effectively recover and enhance underwater images to improve their color, contrast, and sharpness in real-time has become an important and challenging problem in underwater imaging technologies.

Underwater images enhancement (UIE) can utilize traditional methods [7], [8], [9], [10], [11], [12] and learning-based methods [13], [14], [15], [16], [17], [18], [19]. Traditional UIE techniques typically depend on prior knowledge, assumptions, and design principles for processing underwater images, such as white balance [20] and histogram equalization [8]. While these methods are straightforward to apply, their effectiveness is restricted in the adaptability to varying water conditions and lighting situations, producing over-enhanced or under-enhanced results [49].

Learning-based UIE can be generally classified into convolutional neural networks (CNN) -based, and generative learning -based approaches. CNN-based UIE leverages large amount of data to learn the mapping from the degraded underwater image to its corresponding clear image without degradation, and thus it can be adaptive under different water conditions and achieve more robust performance than traditional methods. In the generative learning approach, generative adversarial networks (GAN) -based UIE can learn the conversion between degraded image distribution and the degradation-free image distribution and it has shown great success [21]. However, GAN-based UIE methods suffer from unstable training and mode collapse, generating samples with a lack of diversity [19], and thus they fail represent the full range of possible enhancements.

Another direction in generative learning is the denoising diffusion probabilistic models (DDPM) [22].

Diffusion models have been established as a new state-of-the-art baseline in image-to-image (I2I) prediction tasks, from image super-resolution [23] to image coloring and image restoration [24]. In general, DDPM consists of two processes: 1) the forward diffusion, and 2) the reverse diffusion. The forward diffusion gradually adds Gaussian noise to the input image until its distribution becomes pure Gaussian noise. The reverse diffusion gradually removes the added noise from the noisy image to recover the original image. The unconditional DDPM is generally used in image generation, generating new images with high quality and high diversity by performing unconditional generation in the reverse diffusion process. However, applying unconditional DDPM to the enhancement task may lead to unwanted enhanced images that do not match the input data distribution (i.e.: different semantic information). The conditional DDPM is generally used for I2I tasks, feeding the input image as a prior condition, and guiding the reverse diffusion to generate the enhanced image with the same level of semantic information as the input image. However, for image enhancement in underwater, using solely conditional DDPM as guidance is insufficient and leads to inadequate enhancement results, possibly due to the limited quality of reference enhanced images used for training [19], since they were obtained from the best-of-all enhanced results by other UIE methods. A previous study UW-DDPM [19] proposed a dual U-Net network applying a modified conditional diffusion process for both degraded underwater image and ground truth enhanced image, to better fit the degraded image distribution to the enhanced image distribution and accomplish sufficient enhancement. Specifically, the authors utilized two identical U-Net models, one originally used for denoising, and the other one used for data distribution transformation. Every step of forward diffusion trains the Data Distribution Transformation U-Net to fit the conversion between the two distributions, while every step of inverse diffusion trains the Denoising U-Net, and the inference process superposes outputs of both U-Nets to obtain a diffused state of the enhanced image. Despite outstanding performance over both traditional and previous learning-based UIE methods, the high computational complexity of these two U-Net models inside UW-DDPM and the long inference time of its Markovian process hinder the real-time ability to enhance underwater images on computational resource constrained devices.

In this paper, the Distribution Transformation U-Net inside UW-DDPM, and its Markovian inference process are further investigated. As a result, a new physical-based and diffusion-based UIE approach, called UW-DiffPhys, is proposed to reduce the computational complexity of UW-DDPM while maintaining comparable performance with the original model, and additionally accelerate the inference process of UW-DDPM. Based on these improvements, the overall goal of the proposal aims at improving real-time performance of the diffusion-based UIE on a low-cost AUV. The effectiveness of the proposal can be evaluated based on UIE metrics, inference time, and computational complexity, for a specific task and hardware constraint of AUV.

The contributions of this paper are as follows:

- To reduce computational complexity while maintaining comparable performance with existing Diffusion-based UIE (UW-DDPM) [19], this paper proposes a novel Diffusion Underwater Physical Model, leveraging light-computation physical-based UIE network components and the support from existing Denoising U-Net, to replace the high-computation Distribution Transformation U-Net.
- To accelerate the inference process while maintaining the same distribution after superposition in every time step, the sampling strategy and the distribution shifting during inference are modified with a deterministic implicit sampling technique.
- Regarding the UIE metrics, the proposed UW-DiffPhys model is evaluated qualitatively and quantitatively with the traditional, CNN-based, GAN-based, and Diffusion-based UIE methods. Experiment results show a slight decrease of the proposal in PSNR, SSIM, UCIQUE metrics, but a considerable increase in UIQM performance compared to UW-DDPM, while reducing the computational complexity to approximately half of UW-DDPM, as well as the total inference time.
- Regarding specific AUV task and hardware constraint, due to the outperformance in UIQM (the overall quality UIE metric) when comparing UW-DiffPhys to UW-DDPM, their performance in feature points matching task is evaluated, which is an essential step for AUV localization application. The potential of operating the diffused-based UIE methods on a low-cost NVIDA Jetson Nano is also discussed.

The rest of the article is organized as follows. Section II reviews the background and related work. In Section III, the proposed UW-DiffPhys is described in detail. In Section IV, extensive experiments are provided to evaluate the effectiveness of the proposal, compared to other baselines. The discussion on the experiment results is given in Section V. Finally, Section VI concludes this article.

## II. Background and Related Work

This section introduces the background of underwater physical imaging model, related studies on underwater image enhancement tasks, and fundamentals of Denoising Diffusion Probabilistic Model.

### *A. Underwater Physical Image Formation Model*

Underwater imaging is influenced by the attenuation effect, which involves factors such as light absorption and scattering [33]. When light travels through water, it gets absorbed differently based on its wavelength, as depicted in Fig. 1. For instance, red light, which has a longer wavelength, is absorbed more quickly, and loses intensity at shallower depths. In contrast, blue and green lights, with shorter wavelengths, penetrate deeper, giving underwater images a bluish or greenish hue. Underwater photographs capture light from two main sources: directly transmitted light and background scattered light. The directly transmitted light comes from the object and retains its original radiance but is weakened by forward-scattering and absorption, leading to color distortion. Forward-scattering slightly affects pixel intensity and is usually negligible [43]. However, the background scattered light caused by ambient light scattering off numerous tiny particles in the water, creates hazy or blurry images with low contrast.

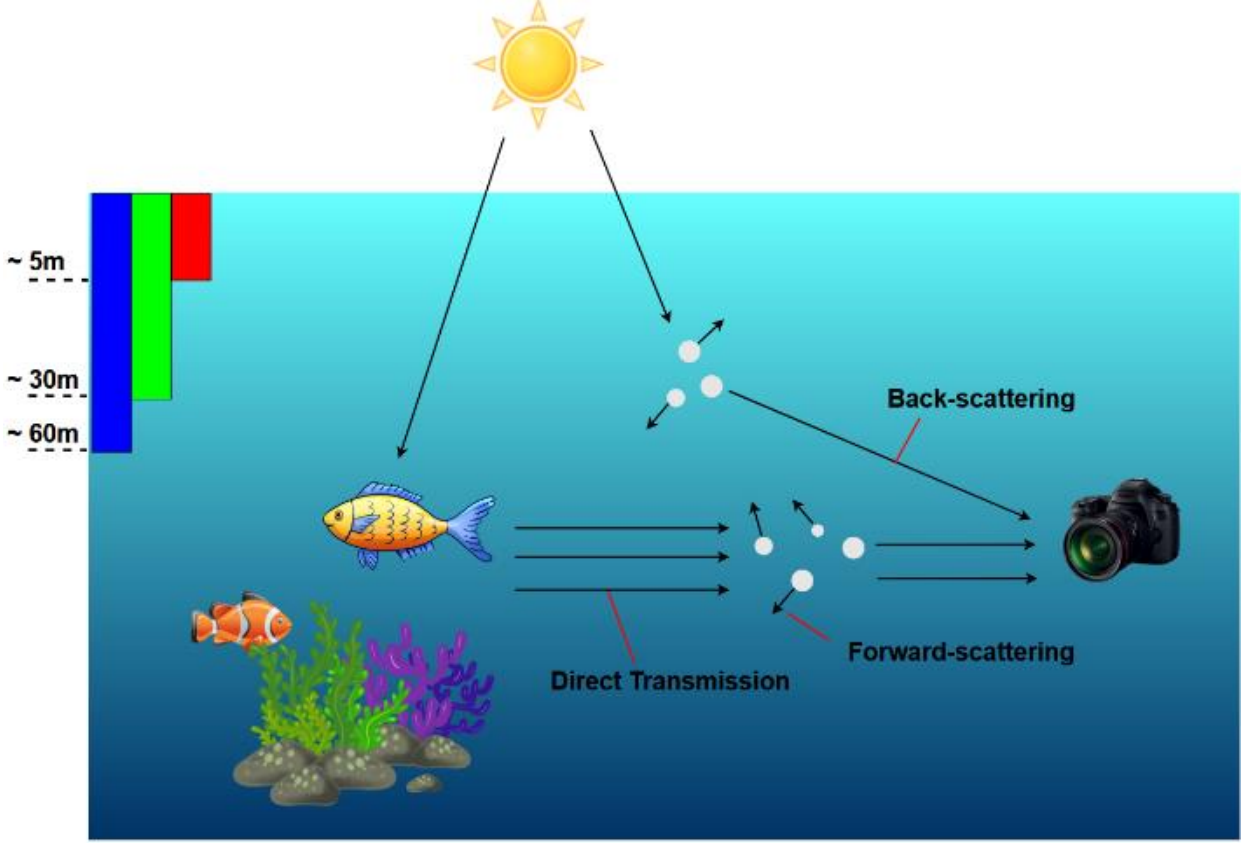

**Figure 1.** Illustration of Underwater Optical Imaging

The mathematical formulation for underwater physical imaging model has been widely obtained from the hazing image model [33]:

$$I(x) = D(x)t(x) + A\left(1 - t(x)\right) \tag{1}$$

Here, $x$ represents the pixel coordinates, $I(x)$ is the raw underwater image affected by various degradations, $D(x)$ is the true radiance of the object (or the restored image after correcting underwater degradations), $t(x)$ is the medium transmission map, and A stands for ambient light. The term $A(1 - t(x))$ represents the backscattered light, while $D(x)t(x)$ represents the directly transmitted light. The true radiance $D(x)$ is attenuated according to the attenuation coefficient $\beta$ and the transmission distance $d$ as represented in the transmission map as follows:

$$t(x) = e^{-\beta d} \tag{2}$$

Given the wavelength-dependent color absorption in water and the unaccounted dependencies in backscattered light from the hazing model, the underwater image model is revised by adjusting the attenuation coefficients [44] as follows:

$$I_c(x) = D_c(x)e^{-\beta_c^D d} + A_c\left(1 - e^{-\beta_c^B d}\right) \tag{3}$$

where $c$ denotes the color channel, $\beta_c^D$ and $\beta_c^B$ are the direct transmission and backscattered attenuation coefficients, respectively. Equation (3) can be reorganized to retrieve the object's true radiance:

$$D_c(x) = (I_c(x) - A_c)e^{\beta_c^D d} + A_c e^{(\beta_c^D - \beta_c^B)d} \tag{4}$$

Since $\beta_c^D - \beta_c^B$ is very small and can be disregarded for $d > 3m$ [33], Eq. (4) is simplified as follows:

$$D_c(x) = (I_c(x) - A_c)e^{\beta_c^D d} + A_c \tag{5}$$

### B. *Underwater Image Enhancement*

#### 1) *Traditional UIE Method*

The traditional UIE methods often rely on prior knowledge or assumptions about the environment, or specific design rules to process the underwater images. These image degradation priors are based on physical models to perform the inverse image degradation process. Li et al. [45] proposed an UIE algorithm based on dark channel prior to recover the blue-green channel, while correcting the red channel using the gray-world assumption and carrying out brightness and contrast balancing with adaptive exposure map. Berman et al. [11] proposed underwater image color restoration method using underwater physical imaging model with parameters estimated from haze-lines prior. The pixel values on these haze-lines are used to compute attenuation ratios of blue-green and blue-red color channels, which simplifies UIE into single-channel haze removal problem and achieves contrast enhancement and color calibration. Jin et al. [8] proposed an adaptive histogram transformation method to adjust the curve of transformation function using local mean and variance of gray-level prior to enhance image details and contrast. However, these methods did not take into account the inherent noises and artifacts in underwater images, leading to over-enhancement with color distortions or loss of details. To tackle this problem, Li et al. [46] applied minimum information loss principle along with histogram distribution prior. The method estimates the transmission map by minimizing the information loss, while applying color correction based on natural image histogram distribution prior to restore the clear image. However, this method could only obtain limited contrast enhancement, edge preservation, and noise suppression. To further improve the underwater image quality, [9] proposed a multi-algorithm fusion technique in multiple RGB and HSV color spaces. Drews et al. [10] proposed an UIE method based on underwater dark channel prior (UDCP) to utilize statistical priors from outdoor natural images, while considering the blue-green color channels as the main sources of underwater visual information. Nevertheless, these methods also remain other limitations, such as not taking into account the scattering effect in the underwater attenuation problem, or noise protection and detail preservation during enhancement, leading to color distortions and reduced contrast. To summarize, these traditional UIE methods were designed to tackle different limited aspects in underwater degradation problem, based on specific prior knowledge, and thus they tend to not generalize well for diverse underwater conditions. This generally leads to side effects such as over-enhancement (color or contrast distortion) and loss of details.

#### 2) *Deep learning -based UIE Method*

The continuous development of deep learning in recent years has established learning-based methods as baseline models for UIE tasks. The pure CNN-based UIE methods, which focus on image processing, leverage convolutional and pooling layers to extract and learn multilevel high-level image features from underwater images. Li et al. [13] proposed lightweight UWCNN with an enhancement branch, which is designed specifically for each water type to adapt to diverse underwater scenes. The same authors in [15] also proposed CNN-based UIE model named Ucolor, to learn from multiple color spaces to highlight and integrate their most discriminative features. Sun et al. [47] utilized convolutional layers to filter noise, along with deconvolutional layers to recover details and optimize image. To improve real-time performance of learning-based UIE, Naik et al. [6] proposed Shallow-UWNet architecture with fewer parameters than existing models while maintaining strong enhancement performance. Chen et al. [33] integrates lightweight CNN modules into underwater physical image formation model, by training two network components, one to estimate the backscattered light, the other to estimate the direct transmission. The method enhanced richer details by removing the estimated backscattered light, obtained limited color improvement, and achieved higher PSNR, SSIM scores.

With the rapid advancement of generative learning, generative adversarial networks (GANs) has recently become the starting model for image-to-image tasks, including underwater image enhancement. Zhu et al. [48] proposed WaterGAN learning underwater imaging models from unlabeled underwater video sequences and generate synthetic underwater images with high realism. Liu et al. [16] proposed a conditional MLFc-GAN utilizing multilevel feature fusion to improve the contrast and color of underwater images. The multilevel feature fusion technique integrates local features into global features, hence enhancing the learning ability of the network. Islam et al. [17] also proposed a conditional GAN model named FUniEGAN, incorporating an objective function that comprehensively considers local to global texture, color, and style information to guide the adversarial training process, making it applicable for both paired or unpaired underwater images data. However, these GAN-based approaches suffer from unstable training process and difficult to converge. Additionally, their generated enhanced images often have uncertainty or diversity in texture, causing inconsistency in content, color, and structure with the ideal clear underwater image without degradation.

### *C. Denoising Diffusion Probabilistic Model*

Another direction in generative learning is the denoising diffusion probabilistic models (DDPM) [22], which can tackle the limitations of GAN-based methods and achieve state-of-the-art performance in image-to-image tasks, including UIE. This section first introduces the background on DDPM, then reviews the previous study in the diffusion-based UIE task.

#### *1) Mathematical Background*

Developed from theoretical basis of generative method inspired by nonequilibrium thermodynamics [31], DDPM is a simplified diffusion model, learning input data distribution through forward and inverse diffusion processes. The forward diffusion gradually adds Gaussian noise to the input image until it becomes an isotropic Gaussian distribution. The learning of whole data distribution can be gradually achieved in the inverse process, predicting the small amount of noise added in every step of the forward process, or modeling the small transformation from simple (Gaussian) to complex (original input) distribution.

##### *a) Forward Diffusion Process*

The forward diffusion can be defined as a Markov chain satisfying probability density $q$ . This process continuously adds a predefined amount of Gaussian noise, to the input image distribution through $T$ iterations, which can be derived as follows:

$$q(x_{1:T}|x_0) = \prod_{t=1}^{T} q(x_t|x_{t-1}) \tag{6}$$

$$q(x_t|x_{t-1}) \sim N\left(x_t \middle| \sqrt{\alpha_t} x_{t-1}, (1-\alpha_t)I\right) \tag{7}$$

, where $\alpha_t \in (0,1)$ is a hyperparameter determining noise variance added at each iteration, $x_0$ is the input distribution, and $x_t$ is the noisy distribution of $x_0$ at iteration t.

Given input $x_0$, the noisy distribution iteration t can be calculated from Eq. (6, 7) as follows:

$$q(x_t|x_0) \sim N\left(x_t \middle| \sqrt{\bar{\alpha}_t} x_0, (1-\bar{\alpha}_t)I\right) \tag{8}$$

$$x_t = \sqrt{\bar{\alpha}_t} x_0 + \sqrt{1-\bar{\alpha}_t} z_t, z_t \sim N(0,1) \tag{9}$$

where $\bar{\alpha}_t = \prod_{i=1}^{t} \alpha_i$.

##### *b) Inverse Diffusion Process (Denoising)*

The denoising process starts from an isotropic Gaussian distribution and gradually recovers to the original input distribution through reversed iterations. In a formal definition, this process aims to obtain the posterior distribution $p_\theta(x_0|x_1)$ by iteratively solving previous posterior distribution $p_\theta(x_{t-1}|x_t)$, where $p_\theta$ denotes the distribution predicted by a neural network $f_\theta$.

After sampling from a standard Gaussian distribution $x_T \sim N(0, I)$, the posterior distribution of $x_{0:T}$ can be defined as follows:

$$p_\theta(x_{0:T}) = p(x_T) \prod_{t=1}^{T} p_\theta(x_{t-1}|x_t) \tag{10}$$

$$p_\theta(x_{t-1}|x_t) \sim N(x_{t-1};\ \mu_t, \sigma_t^2 I) \tag{11}$$

, where $p(x_{t-1}|x_t, x_0)$ denotes the posterior distribution of $x_{t-1}$ given $x_t$ and $x_0$ , whereas $\mu_\theta, \sigma_\theta^2$ represent the mean and variance respectively.

$$\mu_t = \frac{\sqrt{\bar{\alpha}_{t-1}}(1-\alpha_t)}{1-\bar{\alpha}_{t-1}} x_0 + \frac{\sqrt{\alpha_t}(1-\bar{\alpha}_{t-1})}{1-\bar{\alpha}_t} x_t \tag{12}$$

$$\sigma_t^2 = \frac{(1-\bar{\alpha}_{t-1})(1-\alpha_t)}{1-\bar{\alpha}_t} \tag{13}$$

Eq. (12, 13) shows that the variance is known based on selected hyperparameter, and the mean is unknown quantity depending on $x_0$ and $x_t$. However, combining Eq. (9) into Eq. (12), the mean can depend only on $x_t$ as follows:

$$\mu_t = \frac{1}{\sqrt{\alpha_t}} \left( x_t - \frac{1-\alpha_t}{\sqrt{1-\bar{\alpha}_t}} z_t \right) \tag{14}$$

Therefore, the random noise $z_t$, which was previously sampled at forward diffusion step $t$, can be predicted by a neural network $f_\theta(x_t, t)$, from which the mean of posterior distribution $p_\theta(x_{t-1}|x_t)$ can be calculated. The network repeats the process for T iterations, and eventually obtains a pseudo image satisfying input image distribution.

*c) Deterministic Implicit Sampling (DDIM)*

Since the original DDPM [22] requires the total number of time steps T to be sufficiently large (i.e., T=1000) to generate high quality sample, the execution time for inverse diffusion also increases correspondingly. Denoising diffusion implicit model (DDIM) [32] was proposed to accelerate sampling process for pre-trained DDPM models by a deterministic approach. DDIM exploits a generalized formulation of non-Markovian process:

$$q_\lambda(x_{1:T}|x_0) = q_\lambda(x_T|x_0)\prod_{t=2}^{T} q_\lambda(x_{t-1}|x_t, x_0) \quad (15)$$

$$q_\lambda(x_{t-1}|x_t, x_0) \sim N(x_{t-1};\ \tilde{\mu}_t, \lambda_t^2 I) \quad (16)$$

$$\tilde{\mu}_t = \sqrt{\bar{\alpha}_{t-1}}\, x_0 + \sqrt{1 - \bar{\alpha}_{t-1} - \lambda_t^2}\, z_t \quad (17)$$

When setting $\lambda_t^2 = \sigma_t^2$ from Eq. (13), the process becomes Markovian and similar to the original DDPM. Importantly, DDIM utilizes this setting to achieve the same training objective, but only applies non-Markovian formulation with different $\lambda_t^2$ setting for sampling or inverse diffusion process.

A deterministic implicit sampling approach applies $\lambda_t^2 = 0$, and the mean can be derived as follows:

$$\tilde{\mu}_t = \sqrt{\bar{\alpha}_{t-1}}\left(\frac{x_t - \sqrt{1 - \bar{\alpha}_t}z_t}{\sqrt{\bar{\alpha}_t}}\right) + \sqrt{1 - \bar{\alpha}_{t-1}}z_t \quad (18)$$

, with $z_t = f_\theta(x_t, t)$ also predicted by a denoising network.

To accelerate sampling, one can select a sub-sequence $\{\tau_1, \tau_2, \dots, \tau_S\}$ from the complete {1, …, T} time steps. This would utilize the non-Markovian process $q_\lambda(x_{t-1}|x_k, x_0)$, where $k \geq t$. The sub-sequence can be chosen by uniformly interleaving from {1, …, T}:

$$\tau_i = \frac{(i-1)T}{S} + 1 \quad (19)$$

, setting $\tau_1$=1 as the final sampling step.

*d) Conditional Diffusion Models*

The conditional diffusion models aim to learn conditional posterior distribution $p_\theta(x_{0:T} \mid \tilde{x})$ in the inverse diffusion process, instead of $p_\theta(x_{0:T})$ as in DDPM, so that the sampled data has high fidelity to the distribution of $\tilde{x}$.

During training, a paired data distribution (i.e., raw underwater image $x_0$ and its reference image $y_0$ without degradation) is sampled $(x_0, y_0) \sim q(x_0, y_0)$, and the conditional diffusion model is learnt with $y_0$ as guidance in the inverse process:

$$p_\theta(x_{0:T} \mid y_0) = p(x_T)\prod_{t=1}^{T} p_\theta(x_{t-1}|x_t, y_0) \quad (20)$$

Thus, the denoising network $f_\theta(x_t, y_0, t)$ receives the additional $y_0$ as input. For image-to-image (I2I) tasks, $x_t$ and $y_0$ are usually concatenated channel-wise.

*2) Denoising Diffusion Probabilistic Model -based Underwater Image Enhancement*

The rapid development of diffusion models has resulted in an increasing number of studies in the field of image generation [25],[26],[27], and image enhancement [28],[29],[30] using diffusion and conditional diffusion approaches. Regarding image enhancement in the underwater domain, since solely applying a conditional diffusion approach does not sufficiently enhance the image, Lu et al. [19] proposed a modified conditional diffusion UW-DDPM model, and employed a dual U-Net networks for image denoising task and image distribution transformation task to effectively recover the distribution of images without degradation from raw underwater images. The conditional diffusion process is modified to emphasize more on the contribution of its adaptive guidance as follows. UW-DDPM applies the conditional diffusion on the reference image $y_0$ instead of the raw underwater image $x_0$, while the guidance adapted for each diffusion step utilizes the predicted transformation $y'_t = f_\phi(x_t, t)$ from the diffused raw underwater image $x_t$ to the diffused reference image $y_t$. Compared to Eq. (20), the modified conditional diffusion model is learnt with $y'_t$ as guidance in the inverse process as follows:

$$p_\theta(y_{0:T} \mid y'_t) = p(y_T)\prod_{t=1}^{T} p_\theta(y_{t-1}|y_t, y'_t) \quad (21)$$

Beside using a heavy-computation U-Net for the denoising network $f_\theta$, the transformation $f_\phi$ from the raw to the reference image at every diffused state is also obtained by another U-Net in their work [19]. Additionally, a high number of sampling steps (i.e., T=1000) is still required for the inference process due to the use of Markovian DDPM as in Eq. (10), leading to a significant computation time processing each image or video frame, and hindering the real time performance of downstream AUV applications.

This section has introduced the background of underwater physical image formation model supporting the general UIE problem formulation, and how the UIE methods had evolved from the traditional to deep learning -based approaches. Within the deep-learning methods, the transition from CNN-based to generative learning -based methods has also increased in recent years. Within the generative learning -based techniques, this section also highlighted DDPM as an emerging solution for image-to-image tasks, including UIE, tackling limitations of existing GAN-based methods. Nevertheless, the existing diffusion-based UIE [19] did not tackle the slow sampling speed of the diffusion model in the inference process, or further investigate the motivation behind using another U-Net for the distribution transformation $f_\phi$ task, leading to the use of high computational resources when applying this solution. In the next section, the distribution transformation network and the inference process are further investigated and enhanced, aiming at improving real-time performance of the diffusion-based UIE method on a low-cost AUV.

## III. METHODOLOGY

This section first provides an overview of the proposed UW-DiffPhys model. To address the issue of high computational distribution network, a novel Diffusion Underwater Physical Model is then proposed in details in the training process, leveraging light -computation physical-based UIE network components and the support from existing Denoising U-Net. For the inference process, to reduce the required time steps and accelerate the total inference time, the sampling strategy and the distribution shifting during inference are modified with a deterministic implicit sampling technique.

### A. Overview

Unlike the inverse diffusion process of DDPM, both the proposed UW-DiffPhys and UW-DDPM[19] are conditional diffusion models since the UIE task aims at transforming the distribution of degraded images into the distribution of reference images without degradation. This can be interpreted as solving the conditional probability $p(y|x)$ , in which x denotes the raw underwater image and $y$ denotes its reference image. Therefore, the raw image $x$ is taken as a priori condition to guide the inverse diffusion process, generating the enhanced image $y'$ with the same semantic level as $x$. Nevertheless, previous study [19] demonstrated that a single conditional Denoising Network is not sufficient to complete the transformation of data distribution. Thus, a modified conditional diffusion process is required to obtain adaptive guidance, as shown in Eq. (21). This requires a combination of Denoising Network and Distribution Transformation Network for the diffusion-based UIE task. However, in contrast to UW-DDPM [19] using dual high-computation U-Net models for both networks, the proposed UW-DiffPhys combines light-computation physical-based UIE network components with existing Denoising U-Net to replace the high-computation Distribution Transformation U-Net.

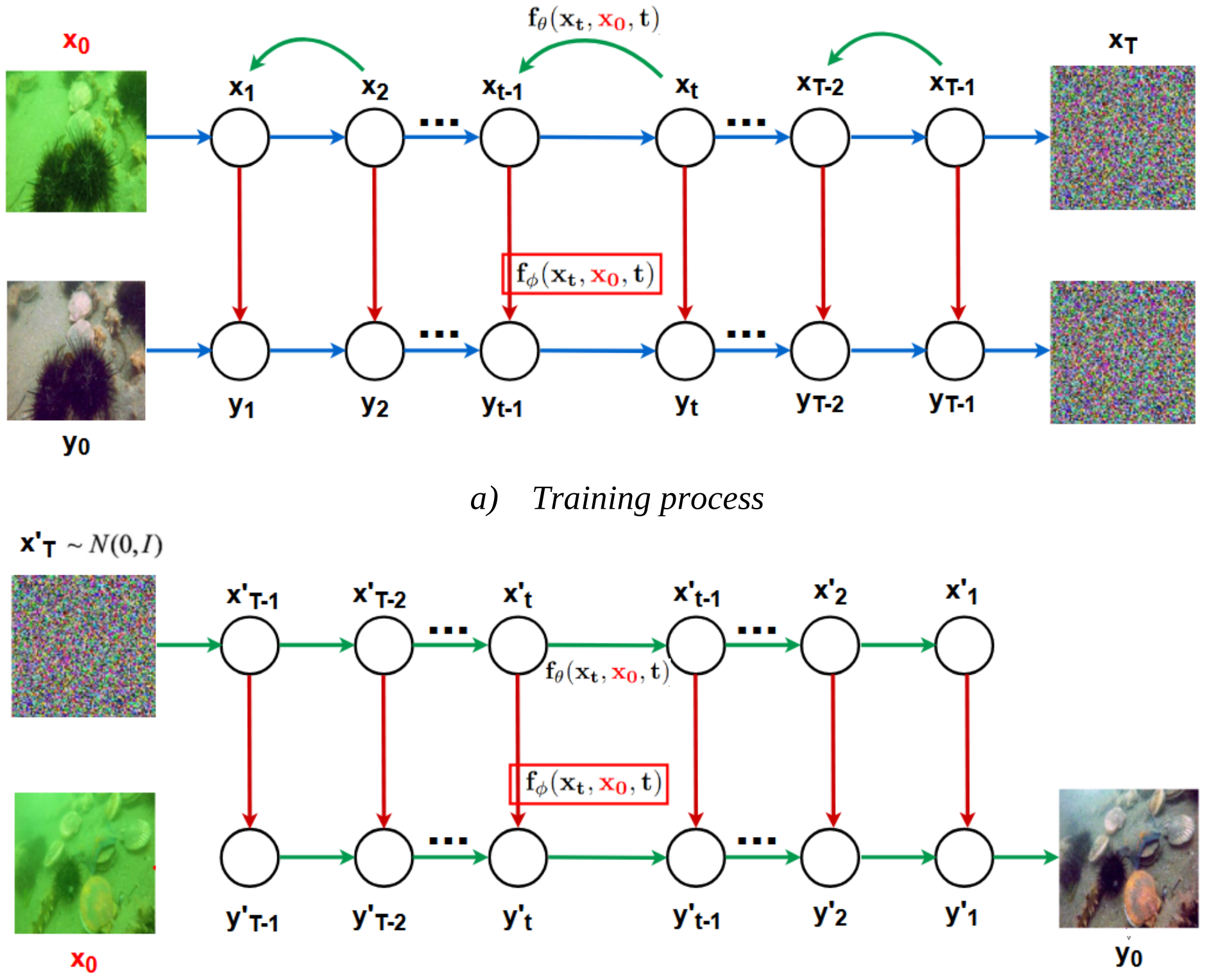

*a) Training process*

*b) Inference process*

**Figure 2.** Overview training and inference processes of the proposed UW-DiffPhys and UW-DDPM [19]. The blue, green, and red arrows indicate forward diffusion, inverse diffusion (denoising with network $f_\theta$), and distribution transformation (with network $f_\phi$), respectively. The key novelties of UW-DiffPhys are inside the Distribution Transformation Network $f_\phi$, and the accelerated inference process.

UW-DiffPhys can be divided into training and inference procedures, as shown in Fig. 2. $x_0, y_0$ are the original degraded underwater image and its corresponding non-degraded reference image, whereas $x_t$,$y_t$ are their diffused states at time step $t$ respectively. Both procedures utilize the Denoising Network $f_\theta(x_t, x_0, t)$ and Distribution Transformation Network $f_\phi(x_t, x_0, t)$. The conditional Denoising Network (from here referred to as the $\theta$ network) aims to predict the random noise $z_t$ that was sampled at the forward diffusion step $t$, as in Eq. (14), but conditioned on $x_0$, the raw underwater image. The Distribution Transformation Network (from here referred to as the $\phi$ network) aims to convert $q(x_t|x_0)$ into $q(y_t|y_0)$ at each time step. The main novelty of UW-DiffPhys compared to UW-DDPM [19] lies in the $\phi$ network components, which utilize support from existing $\theta$ network, and underlying underwater physical imaging properties from given diffused states $x_t$ and $y_t$. The additional improvement is in the accelerated inference process using a non-Markovian approach. These novelties will be detailed in the description of training and inference processes separately.

### B. Training Process

The training process of UW-DiffPhys involves joint training $\phi$ and $\theta$ networks following forward and inverse diffusion processes, respectively. The input training data can be formalized as $\{x_0, y_0\} = \{x^i, y^i\}_{i=1}^{n}$, in which $x_0$ denotes the degraded underwater image, with its non-degraded reference image $y_0$, and $n$ is the dataset size. The diffused states $x_t, y_t$ at time step t are obtained from Eq. (9) as follows:

$$x_t = \sqrt{\bar{\alpha}_t} x_0 + \sqrt{1 - \bar{\alpha}_t} z_t, z_t \sim N(0,1) \quad (22)$$

$$y_t = \sqrt{\bar{\alpha}_t} y_0 + \sqrt{1 - \bar{\alpha}_t} z_t, z_t \sim N(0,1) \quad (23)$$

It should be noted that the same noise $z_t$ is added to both $x_t$ and $y_t$.

#### 1) Denoising Network θ

In an inverse diffusion step, from an input diffused state $x_t$, the conditional Denoising Network attempts to approximate $p_\theta(x_{t-1}|x_t, x_0)$ by optimizing parameters when estimating the random noise $z_t$ embedded in $x_t$, and conditioned on $x_0$. The loss function of $\theta$ network can be given as follows:

$$Loss_\theta = \mathbb{E}_{x_0, z_t, t}[\|z_t - f_\theta(x_t, x_0, t)\|_2^2] \quad (24)$$

For every time step in the inverse diffusion process, the network concatenates 6-channel $(x_t, x_0)$ as input then outputs the predicted noise $z'_t$, and has a U-Net architecture utilized from [29]. This implementation uses sinusoidal positional encoding for time step embedding, self-attention blocks at 16x16 feature map. Further configuration details on denoising model architecture can be found in Table I.

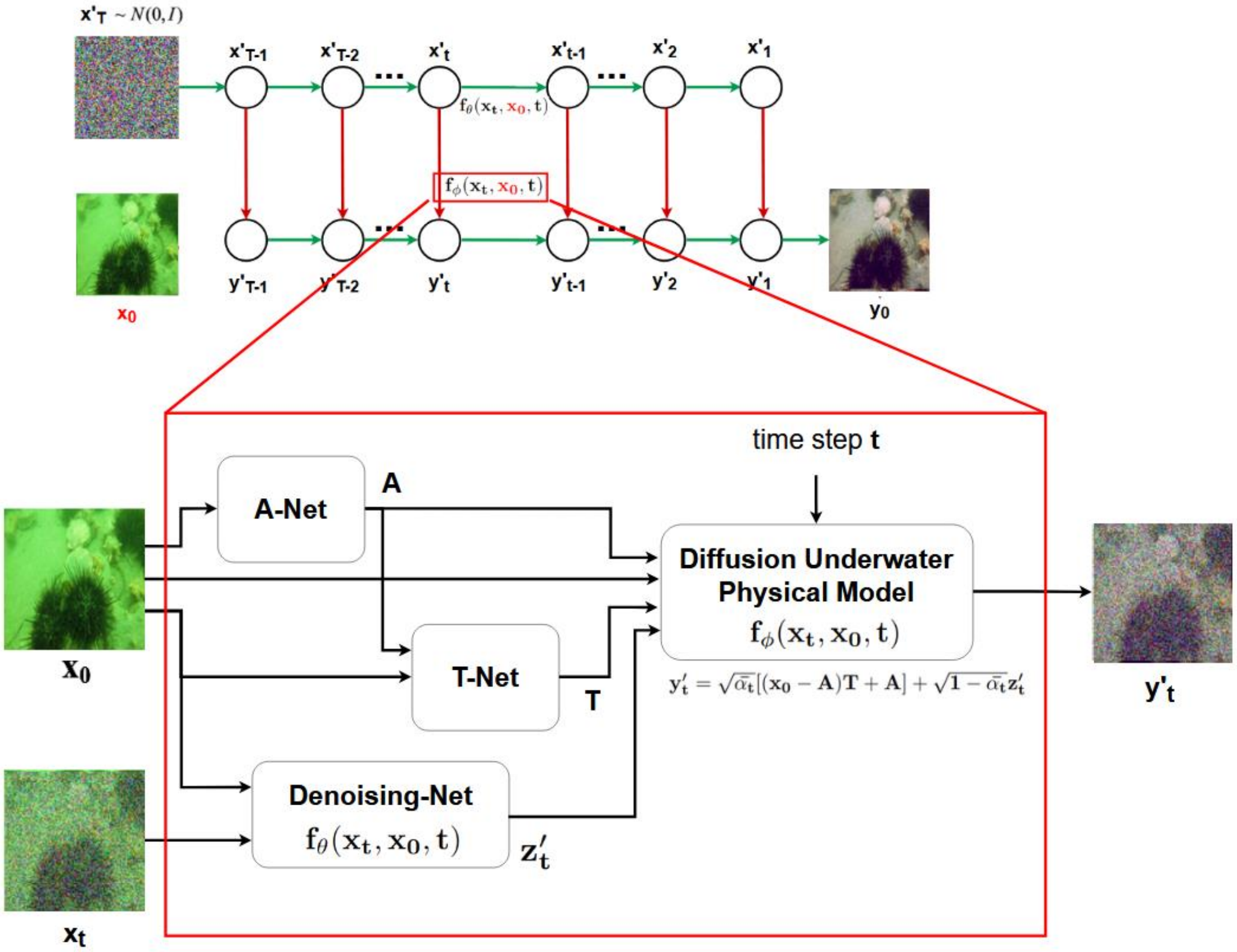

**Figure 3.** Overview of Proposed Diffusion Underwater Physical Model (Distribution Transformation Network $f_\phi$)

#### 2) Distribution Transform Network ϕ

In a forward diffusion step, the Distribution Transformation Network attempts to predict the non-degraded diffused state $y_t$ from the degraded diffused state $x_t$. Since the Transformation task also handles the same noisy image input $x_t$ as Denoising task, UW-DDPM utilizes the same heavy-computation U-Net architecture for the Transformation task. However, as shown in Eq. (22) and (23), $x_t$ and $y_t$ share the same underlying added Gaussian noise $z_t$, which is different from the Denoising task where $x_{t-1}$ and $x_t$ are added with different noise amounts. Therefore, the Distribution Transformation task can be essentially reduced to predicting original non-degraded image $y_0$ from the original degraded image $x_0$, then added with previously predicted noise $z_t$ by the Denoising network. Nevertheless, the network predicting $y_0$ from $x_0$ is essentially a general learning-based UIE model. With the goal of reducing the computational cost, light-computation learning-based UIE models combined with underwater physical imaging properties, such as [33][34], are considered for integration with diffusion model.

*a) Diffusion Underwater Physical Model*

From Eq. (5) and (23), the integration of underwater physical imaging formation into the diffusion context can be formalized as a Diffusion Underwater Physical Model as illustrated in Fig. 3 and expressed as follows:

$$y'_t = \sqrt{\bar{\alpha}_t}[(x_0 - A)T + A] + \sqrt{1-\bar{\alpha}_t} z'_t \quad (25)$$

, where $y'_t$ and $z'_t$ are the predicted non-degraded diffused $y_t$ by $\phi$ network, and the predicted noise $z_t$ by $\theta$ network respectively. A and T are the ambient light and inverse direct transmission maps, estimated by light-computation A-Net and T-Net models. Importantly, instead of using the original Gaussian noise $z_t$ added to both $x_t$ and $y_t$, the noise $z'_t$ estimated by the $\theta$ network is used in Eq. (25). This altered version of $z'_t$ is no longer expected to be pure Gaussian noise and may contain additional information relating to the converting $x_0$ to $y_0$ task. This means the $\theta$ network is simultaneously trained for both the Denoising task, as in Eq. (24), and Distribution Transformation task, as in Eq. (25). In other words, fine-tuning the high-computation Denoising U-Net can compensate for the limited ability of light-computation A-Net, T-Net. Fig. 4 illustrates this concept.

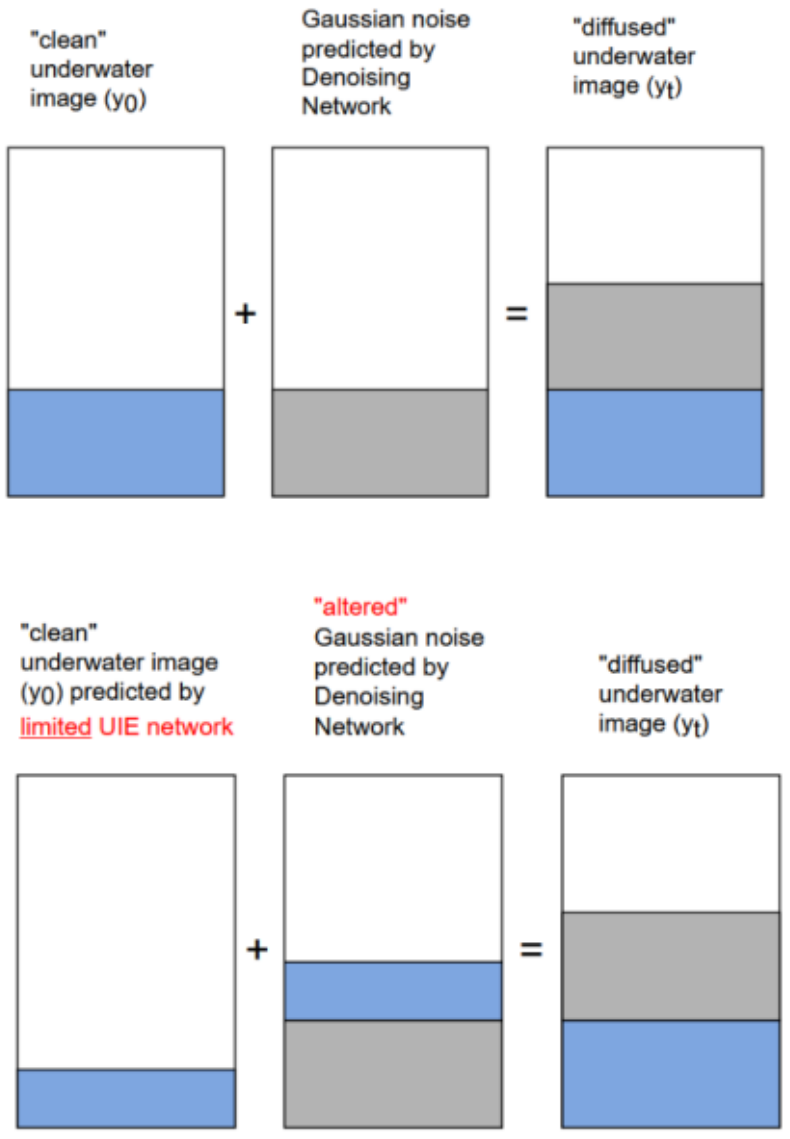

**Figure 4.** Illustration of Denoising Network predicting an "altered" Gaussian noise to compensate for the limited UIE network

*b) A-Net and T-Net*

The A-Net ($f_A$) module estimates the ambient light $A_c$ in Eq. (5), while the T-Net ($f_T$) module estimates the inverse transmission map $e^{\beta_c^D d}$ in Eq. (5). Their network architectures are adopted from [33], as shown in Table II and Table III. The loss function of these modules is defined as follows:

$$Loss_{A,T} = \mathbb{E}_{x_0,y_0}\left[\left\|y_0 - [(x_0 - f_A(x_0) f_T(x_0, f_A(x_0)) + f_A(x_0)]\right\|_2^2\right] \quad (26)$$

*c) Training strategy*

From Eq. (25), the distribution transformation loss can be expressed as follows:

$$Loss_{\phi} = \mathbb{E}_{x_0,x_t,y_t,t}\left[\left\|y_t - y'_t\right\|_2^2\right] = \mathbb{E}_{x_0,x_t,y_t,t}\left[\left\| y_t - \sqrt{\bar{\alpha}_t}[(x_0 - f_A(x_0) f_T(x_0, f_A(x_0)) + f_A(x_0)] - \sqrt{1-\bar{\alpha}_t} f_\theta(x_t, x_0, t) \right\|_2^2\right] \quad (27)$$

The total loss of distribution loss is as follows:

$$Loss_{\phi_total} = Loss_{A,T} + Loss_{\phi}$$

Algorithm 1 illustrates the pseudocode of UW-DiffPhys training process:

**Algorithm 1:** Training denoising model $f_\theta$, and Distribution Transformation model $f_\phi$ in UW-DiffPhys

**Require:** $\chi = \{x^i, y^i\}_{i=1}^n, \alpha, T$

**repeat**

$\{x_0, y_0\} \in \chi$

$z_t \in N(0, I), t \in p(T)$

**Take a gradient descend step on**

$$\nabla_\theta \left\| z_t - f_\theta\left(\sqrt{\bar{\alpha}_t} x_0 + \sqrt{1 - \bar{\alpha}_t} z_t, x_0, t\right) \right\|_2^2$$

$$\nabla_{\phi_total} \left\| \left\| y_0 - \left[ (x_0 - f_A(x_0) f_T(x_0, f_A(x_0)) + f_A(x_0) \right] \right\|_2^2 + \left\| \begin{matrix} y_t - \sqrt{\bar{\alpha}_t}\left[ (x_0 - f_A(x_0) f_T(x_0, f_A(x_0)) + f_A(x_0) \right] \\ - \sqrt{1 - \bar{\alpha}_t} f_\theta(x_t, x_0, t) \end{matrix} \right\| \right\|_2^2$$

**until converged**

TABLE I. U-NET ARCHITECTURE DETAILS OF DENOISING NETWORK

| Path | Layer / Block | Details |
|---|---|---|
| | Input | 6 channels, concatenating 128x128 $x_t$ and $x_0$ |
| | Timestep Embedding | Linear layers: [128 → 512, 512 -> 512] |
| Downsampling | Initial Conv | Conv2d(in_channels=3, out_channels=128, kernel_size=3, stride=1, padding=1) |
| | Downsample - Level 1 | ResnetBlock(in_channels=128, out_channels=128, temb_channels=512, dropout=0.0) |
| | | ResnetBlock(in_channels=128, out_channels=128, temb_channels=512, dropout=0.0) |
| | Downsample - Level 2 | ResnetBlock(in_channels=128, out_channels=256, temb_channels=512, dropout=0.0) |
| | | ResnetBlock(in_channels=256, out_channels=256, temb_channels=512, dropout=0.0) |
| | Downsample - Level 3 | ResnetBlock(in_channels=256, out_channels=384, temb_channels=512, dropout=0.0) |
| | | ResnetBlock(in_channels=384, out_channels=384, temb_channels=512, dropout=0.0) |
| | Downsample - Level 4 | ResnetBlock(in_channels=384, out_channels=512, temb_channels=512, dropout=0.0) |
| | | ResnetBlock(in_channels=512, out_channels=512, temb_channels=512, dropout=0.0) |
| | Downsample - Level 5 | ResnetBlock(in_channels=512, out_channels=512, temb_channels=512, dropout=0.0) |
| | | ResnetBlock(in_channels=512, out_channels=512, temb_channels=512, dropout=0.0) |
| Bottleneck | Resnet Block 1 | ResnetBlock(in_channels=512, out_channels=512, temb_channels=512, dropout=0.0) |
| | Attention Block 1 | AttentionBlock(in_channels=512) |
| | Resnet Block 2 | ResnetBlock(in_channels=512, out_channels=512, temb_channels=512, dropout=0.0) |
| Upsampling | Upsampling - Level 5 | ResnetBlock(in_channels=1024, out_channels=512, temb_channels=512, dropout=0.0) |
| | | ResnetBlock(in_channels=512, out_channels=512, temb_channels=512, dropout=0.0) |
| | Upsampling - Level 4 | ResnetBlock(in_channels=1024, out_channels=384, temb_channels=512, dropout=0.0) |
| | | ResnetBlock(in_channels=384, out_channels=384, temb_channels=512, dropout=0.0) |
| | Upsampling - Level 3 | ResnetBlock(in_channels=768, out_channels=256, temb_channels=512, dropout=0.0) |
| | | ResnetBlock(in_channels=256, out_channels=256, temb_channels=512, dropout=0.0) |
| | Upsampling - Level 2 | ResnetBlock(in_channels=512, out_channels=128, temb_channels=512, dropout=0.0) |
| | | ResnetBlock(in_channels=128, out_channels=128, temb_channels=512, dropout=0.0) |
| | Upsampling - Level 1 | ResnetBlock(in_channels=256, out_channels=128, temb_channels=512, dropout=0.0) |
| | | ResnetBlock(in_channels=128, out_channels=128, temb_channels=512, dropout=0.0) |
| | Output | Normalize(128) |
| | Final Conv | Conv2d(in_channels=128, out_channels=3, kernel_size=3, stride=1, padding=1) |

TABLE II. A-NET ARCHITECTURE DETAILS

| Layer | # Filters | Filter Size | Stride | Padding | Dilation |
|---|---|---|---|---|---|
| Convolution + PReLU | 3 | 3x3 | 1 | 1 | 1 |
| Convolution + PReLU | 3 | 3x3 | 1 | 1 | 1 |
| AdaptiveAvgPool2D | - | - | - | - | - |
| Convolution + PReLU | 3 | 1x1 | 1 | 0 | 1 |
| Convolution + PReLU | 1 | 1x1 | 1 | 0 | 1 |

TABLE III. T-NET ARCHITECTURE DETAILS

| Layer | # Filters | Filter Size | Stride | Padding | Dilation |
|---|---|---|---|---|---|
| Convolution + PReLU | 8 | 3x3 | 1 | 1 | 1 |
| Convolution + PReLU | 8 | 3x3 | 1 | 2 | 2 |
| Convolution + PReLU | 8 | 3x3 | 1 | 5 | 5 |
| Convolution + PReLU | 1 | 3x3 | 1 | 1 | 1 |

### C. *Accelerated Inference Process*

The inference process of general DDPM starts from an isotropic Gaussian noise and gradually iterates back to $x_0$ after T steps. In previous conditional UW-DDPM [19], to obtain sufficiently enhanced output image, the reverse iterations in its inference process are guided by original degraded $x_0$ and the Distribution Transformation network, then eventually obtains $y_0$, as shown in Fig. 2b. Nevertheless, iterating through all T steps would result in high inference time, since T is usually large (i.e., T=1000) to generate high quality output. Therefore, the proposed UW-DiffPhys implements deterministic implicit sampling technique from DDIM [32] models to iterate through only a subset of T steps, and accelerate the inference process.

In each step of the inference process, the output of the Distribution Transformation Network $f_\phi(x_t, x_0, t)$ is summed with the estimated non-degraded state from $f_\theta(y_t, x_0, t)$ to ensure sufficient enhancement and avoid randomness during neural network training [19]. Eq. (22) and (23) show that both outputs of $f_\phi(x_t, x_0, t)$ and $f_\theta(y_t, x_0, t)$ satisfy the Gaussian distribution $N(\mu_t, \sigma_t^2)$, hence $y'_t$ also satisfy the distribution:

$$y'_t = f_\theta(y_t, x_0, t) + f_\phi(x_t, x_0, t) = N(2\mu_t, 2\sigma_t^2) \quad (28)$$

Nevertheless, this distribution obtained after superposition is different from the previous distribution $N(\mu_t, \sigma_t^2)$ in training process of $y_t$ (without superposition), leading to terrible errors in the final enhanced $y_0$. Thus, a distribution shifting operation is required for each stacking to solve this problem [19], expressed as follows:

$$y'_t = \frac{y'_t - 2\mu_t}{\sqrt{2}} + \mu_t \quad (29)$$

Applying the deterministic implicit sampling technique as in Eq. (18), the mean $\mu_t$ can be obtained as follows:

$$\tilde{\mu}_t = \sqrt{\bar{\alpha}_{t-1}} \left( \frac{x_t - \sqrt{1-\bar{\alpha}_t} f_\theta(x_t, x_0, t)}{\sqrt{\bar{\alpha}_t}} \right) + \sqrt{1-\bar{\alpha}_{t-1}} f_\theta(x_t, x_0, t) \quad (30)$$

In summary, the pseudocode of inference process is illustrated in Algorithm 2.

---

**Algorithm 2:** Inference process in UW-DiffPhys

---

**Require:** $\chi = \{x^i\}_{i=1}^n, T$
$\{x_0\} \epsilon \chi$
$sub - sampling \{\tau_1, \tau_2, \dots, \tau_S\} from\ full\ set \{1, \dots, T\}$
$x'_{\tau_S} \sim N(0, I)$
**for** t = $\tau_S$,…, 1 do
  **if** t = $\tau_S$ then
    $x'_{t-1} = \sqrt{\bar{\alpha}_{t-1}} \left( \frac{x'_t - \sqrt{1-\bar{\alpha}_t} f_\theta(x'_t, x_0, t)}{\sqrt{\bar{\alpha}_t}} \right) + \sqrt{1-\bar{\alpha}_{t-1}} f_\theta(x'_t, x_0, t)$
    $y'_{t-1} = f_\phi(x'_{t-1}, x_0, t)$
  **else if** $t > 1$ **and** $t \neq \tau_S$ **then**
    $x'_{t-1} = \sqrt{\bar{\alpha}_{t-1}} \left( \frac{x'_t - \sqrt{1-\bar{\alpha}_t} f_\theta(x'_t, x_0, t)}{\sqrt{\bar{\alpha}_t}} \right) + \sqrt{1-\bar{\alpha}_{t-1}} f_\theta(x'_t, x_0, t)$
    $y'_{t-1} = \sqrt{\bar{\alpha}_{t-1}} \left( \frac{y'_t - \sqrt{1-\bar{\alpha}_t} f_\theta(x'_t, x_0, t)}{\sqrt{\bar{\alpha}_t}} \right) + \sqrt{1-\bar{\alpha}_{t-1}} f_\theta(x'_t, x_0, t) + f_\phi(x'_{t-1}, x_0, t)$
    $y'_{t-1} = \frac{1}{\sqrt{2}} y'_{t-1} + (1-\sqrt{2})[\sqrt{\bar{\alpha}_{t-1}} \left( \frac{y'_t - \sqrt{1-\bar{\alpha}_t} f_\theta(x'_t, x_0, t)}{\sqrt{\bar{\alpha}_t}} \right) + \sqrt{1-\bar{\alpha}_{t-1}} f_\theta(x'_t, x_0, t)]$
  **else if** $t = 1$ **then**
    $y'_{t-1} = \sqrt{\bar{\alpha}_{t-1}} \left( \frac{y'_t - \sqrt{1-\bar{\alpha}_t} f_\theta(x'_t, x_0, t)}{\sqrt{\bar{\alpha}_t}} \right) + \sqrt{1-\bar{\alpha}_{t-1}} f_\theta(x'_t, x_0, t)$
  **end if**
**end for**
**return** $y'_0$

---

## IV. Performance Evaluation

This section evaluates UW-DiffPhys among baseline methods on a set of popular underwater datasets, to demonstrate that the proposal can effectively improve the quality of degraded underwater images, both qualitatively and quantitatively. Additional computational complexity and inference time are also provided to verify the main contribution of this paper.

### A. *Experimental Methods*

#### 1. *Implementation Details*

The implementation of UW-DiffPhys was carried out with Pytorch 2.1.1, on NVIDIA GeForce RTX 3080. The network was trained and early stopped after 2565000 iterations, with a batch size of 4, Adam optimizer with a learning rate of $2 \times 10^{-5}$ without weight decay. An exponential moving average with a weight of 0.999 was used during updating parameters to facilitate more stable learning [35]. The images are resized into 128x128 before training and inference.

#### 2. *Datasets and Evaluation Metrics*

The proposed UW-DiffPhys was conducted on three public underwater image datasets as follow:

1) The UIEB benchmark dataset [36] for UIE contains 950 real underwater images, among which only 890 have high-quality reference images, while the remaining 60 images are challenging without corresponding satisfactory references. The 890 image pairs with high-quality reference images are separated into 800 pairs as training data UIEB-Train, and 90 pairs UIEB-Test90 for both full-reference and non-reference testing, whereas the challenging 60 images are also used for non-reference testing UIEB-Test60.
2) The recent large-scale LSUI dataset [37] contains 5004 image pairs of more diverse underwater scenes with different water types, illumination conditions, and higher quality target references. We combined all 5004 images with UIEB-Train of 8000 images to create a complete training data LSUI-UIEB-Train.
3) The U45 dataset [38] contains 45 underwater images without references in different scenes, with different degradation types, including low contrast, color contrast, and haze effects. This dataset was selected as testing data U45-Test.

The two full-reference metrics PSNR [39] and SSIM [40], along with the two non-reference metrics UCIQE [41] and UIQM [42] specifically designed for underwater scenes, were utilized for UIEB-Test90, UIEB-Test60, U45-Test evaluation datasets. While UCIQE measures the colorfulness and contrast of underwater images by considering chroma, saturation, and contrast, the UIQM metric provides a more comprehensive evaluation of underwater image quality, and considers multiple factors including colorfulness, sharpness, and contrast.

### B. *Experimental Results*

#### 1. *Qualitative and Quantitative Comparison*

The proposed UW-DiffPhys is compared qualitatively and quantitatively on the three testing datasets, against widely used UIE methods in recent years, including traditional method (UDCP [10]), CNN-based methods (UWCNN [13], Shallow-UWNet [14], UW-PhysCNN [33]), GAN-based methods (MLFcGAN [16], FUnIEGAN [17], UW-GAN [18]) , and Diffusion-based method (UW-DDPM [19]). Fig. 5, 6 and 7 provide qualitative results on diverse water types from testing datasets UIEB-Test90, UIEB-Test60, and U45-Test respectively, whereas Table IV summarizes the quantitative results.

The traditional UDCP fails to color-correct and recover limited details in green-toned (i.e: Fig. 5-a, Fig. 5-a,b), blue -toned (i.e: Fig. 5-d,e, Fig. 6-a,b,c,d) or blue-green -toned (i.e: Fig. 5-b,c) images. It also over-enhanced and introduced color distortions to foggy images (Fig. 7-e,f). Shallow-UWNet and UW-PhysCNN achieved limited color-correction for green -toned images, creating yellow -toned results. Nevertheless, UW-PhysCNN performed considerably better in foggy scenes by recovering more details. MLFcGAN and UWGAN showed severe color distortion in green -toned and foggy images, producing yellow-tinted and pinkish results. Meanwhile, the diffusion-based UW-DDPM and our proposed UW-DiffPhys could achieve stable enhancement in most water types. Although UW-DDPM could bring fuller color improvement to the degraded images, the information of the scenes was blurred due to features loss, or the limited ability to remove scattering effect. On the other hand, images enhanced by UW-DiffPhys not only showed similar color improvement (to some extent) as UW-DDPM, but also recovered more details in foggy regions of a scene, especially at further distances. For challenging low-light scenes in UIE-Test60, only UW-PhysCNN, UW-DDPM, UW-DiffPhys can considerably increase the scene brightness. While UW-DDPM can produce visually satisfactory (no color-tainted) results, UW-PhysCNN and UW-DiffPhys can recover sufficient details. To summarize, these stable enhancement results indicate that UW-DiffPhys partially inherited both color improvement ability from diffusion-based UW-DDPM, and scattering removal (details recovering) from physical-based UW-PhysCNN. Figure 8 enlarges a few comparisons between the three methods (from 128x128 resolution) to illustrate the advantage of UW-DiffPhys, recovering more scene details than UW-DDPM, while having fuller color improvement than UW-PhysCNN.

Table IV provides quantitative comparisons between the UIE methods, and further highlights the effectiveness of UW-DiffPhys, balancing performance between physical-based UW-PhysCNN and diffusion-based UW-DDPM. The diffusion-based UW-DDPM obtained the highest value among other UIE methods for the full-reference PSNR and SSIM metrics, indicating its ability to match the distribution of the non-degraded reference images well. Additionally, the ability of UW-DDPM to bring fuller color enhancement could be shown in achieving the highest UCIQUE scores in both non-reference testing sets UIEB-Test60 and U45-Test, while obtaining suboptimal UCIQUE score on UIEB-Test90 with reference data, in which it was outperformed by UDCP due to its over-enhancement effect. Regarding the non-reference UIQM metric depicting a more overall underwater image quality, considering also sharpness and contrast in addition to colorfulness, the physical-based UW-PhysCNN achieved the highest values for all three testing data, due to its ability to remove scattering using physical imaging model. Nevertheless, its color enhancement ability is limited, which is shown in intermediate UCIQUE scores. On the other hand, even though the proposed UW-DiffPhys obtained both suboptimal UIQM and UCIQUE scores, it outperformed UW-DDPM by approximately 7.87%. 18.66%, 6.15% in UIQM metric on UIEB-Test90, UIEB-Test60, U45-Test respectively. Meanwhile, the proposal also outperformed UW-PhysCNN by 0.78%, 1.89% in UCIQUE score on UIEB-Test90, UIEB-Test60 respectively.

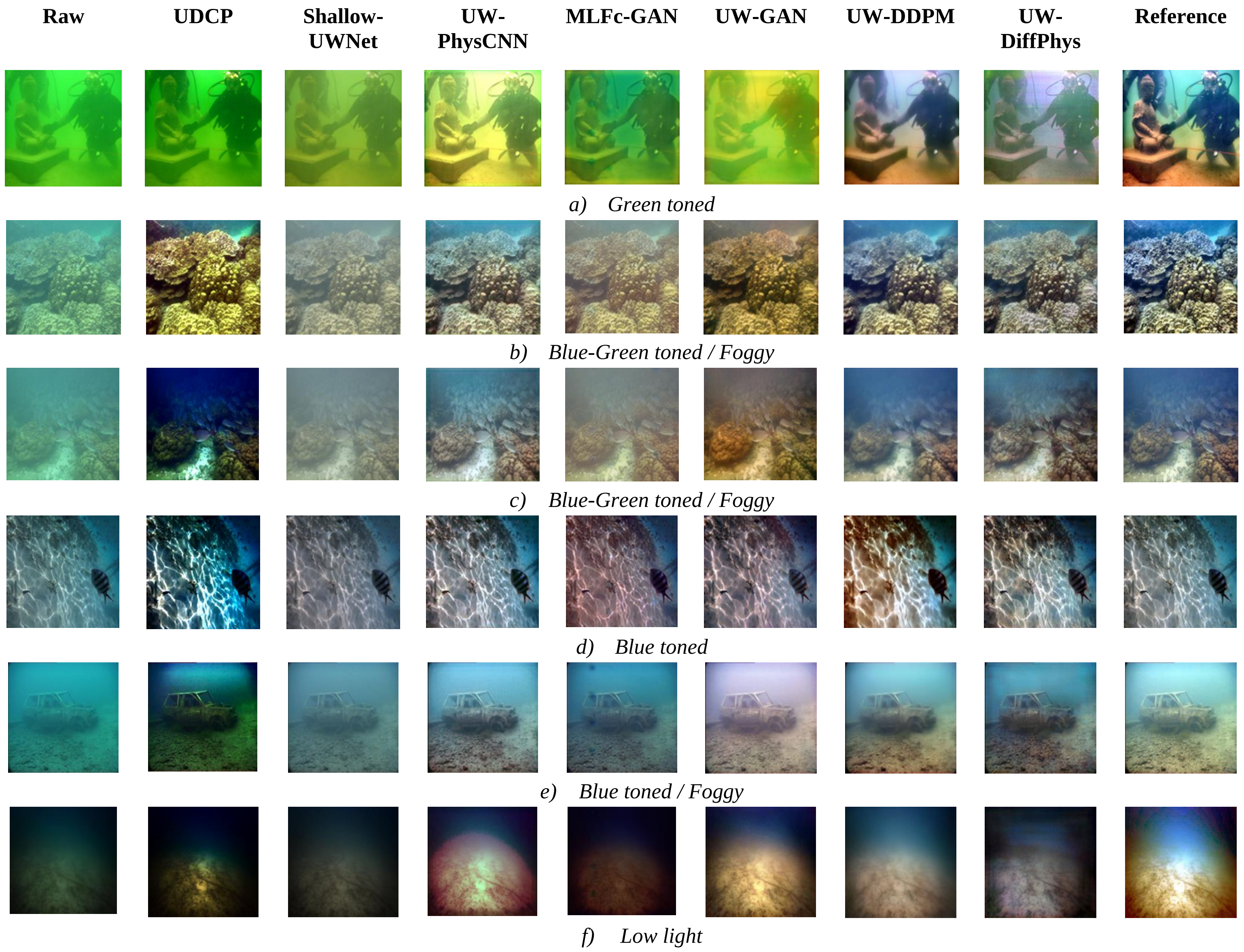

**Figure 5.** Qualitative Comparison between Proposed UW-DiffPhys and Baseline UIE Methods on UIEB-Test90

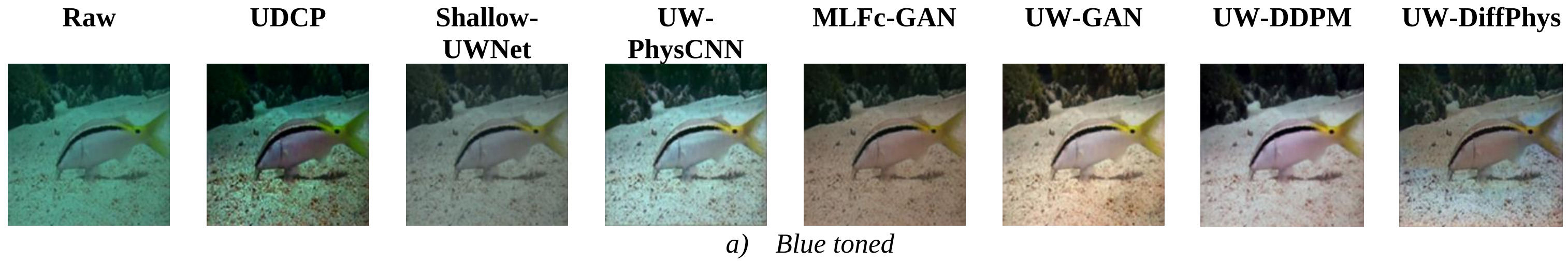

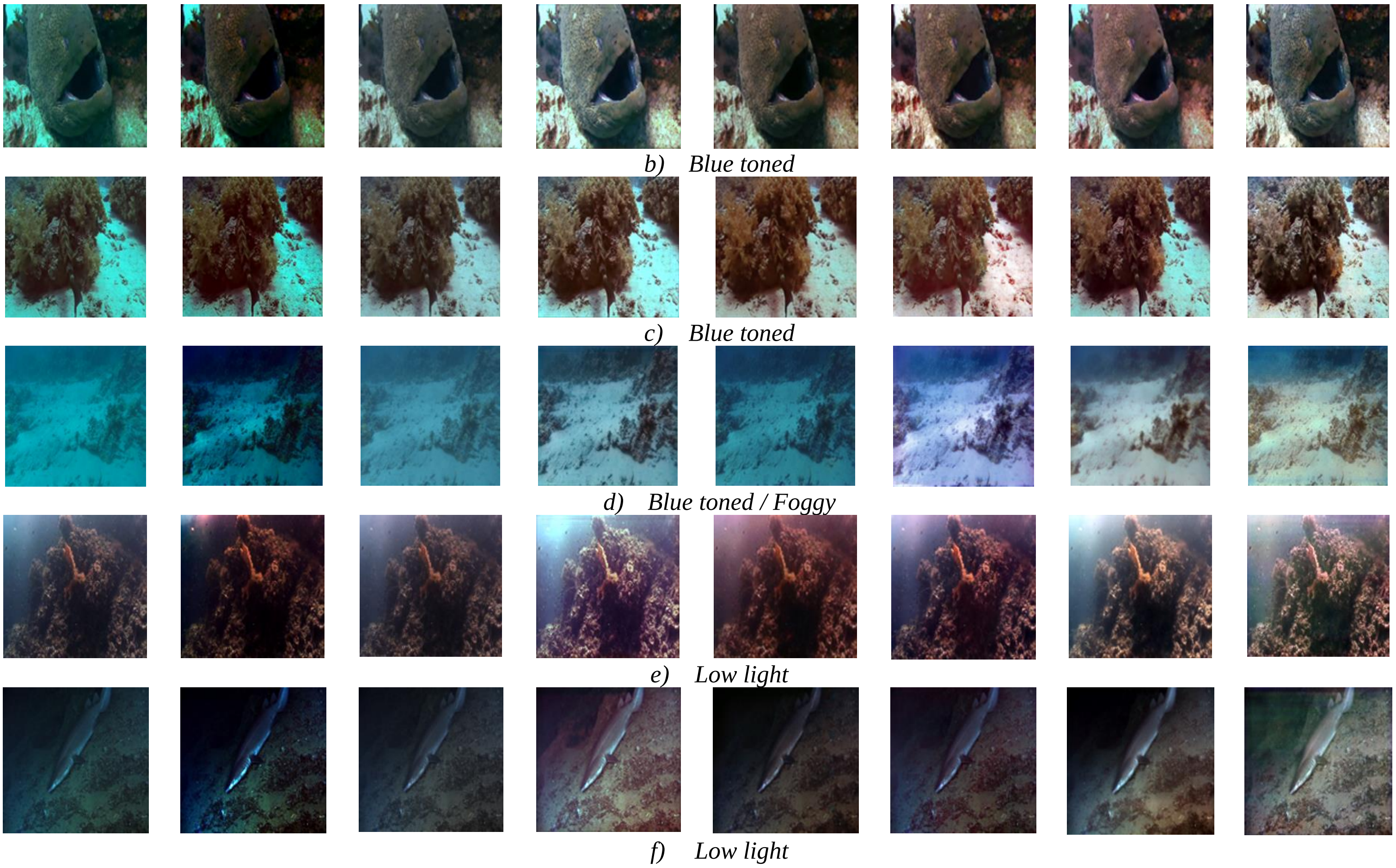

**Figure 6.** Qualitative Comparison between Proposed UW-DiffPhys and Baseline UIE Methods on UIEB-Test60

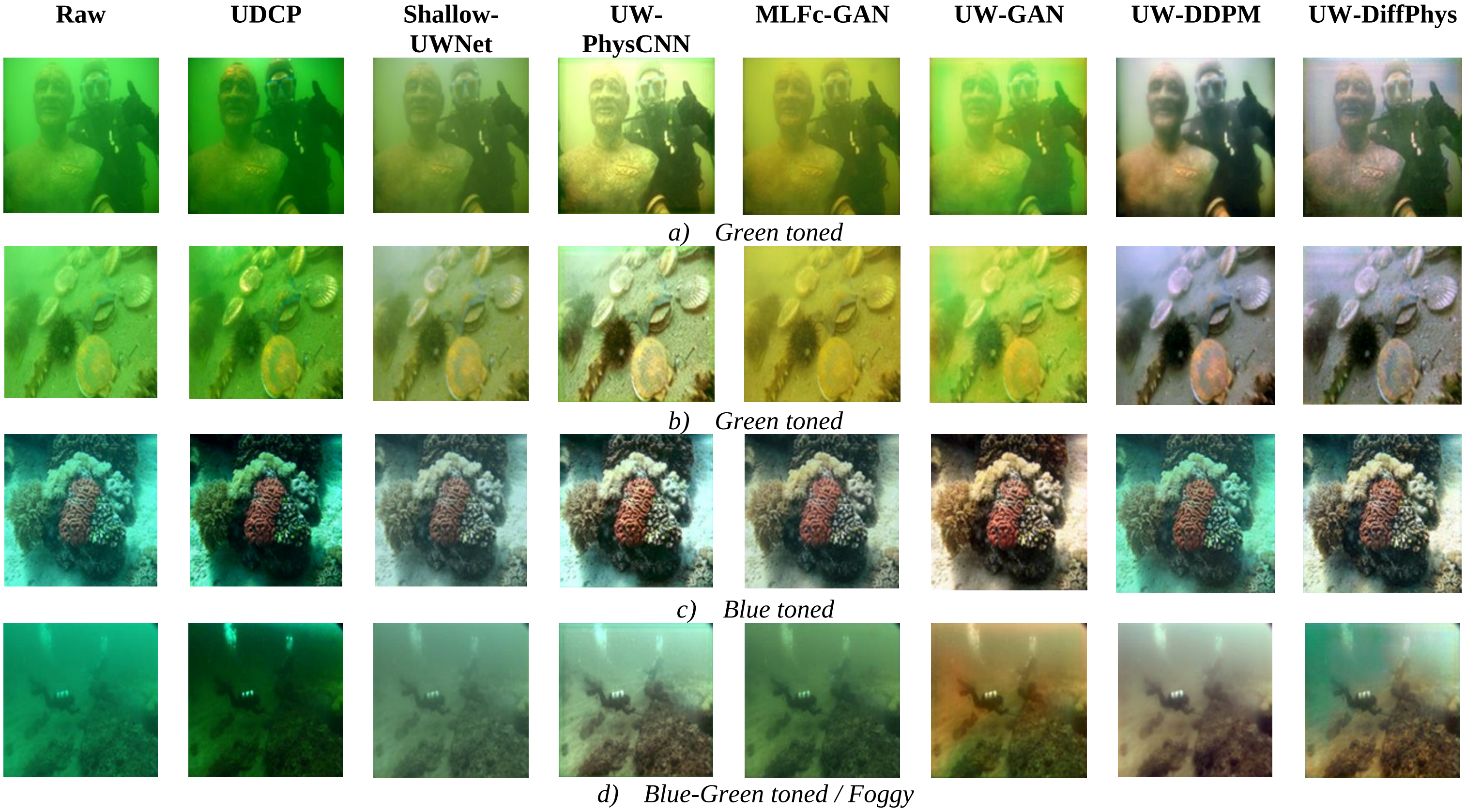

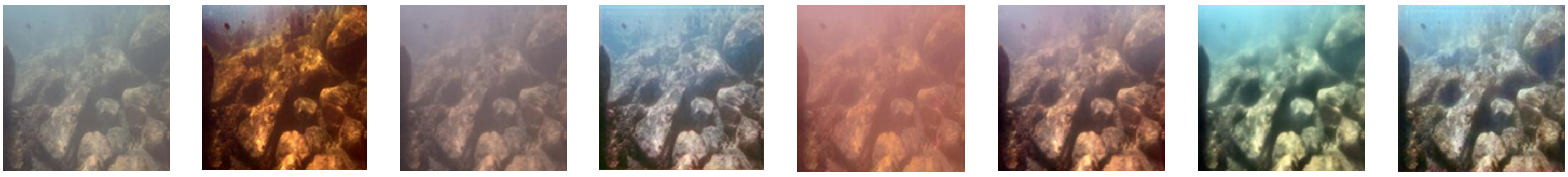

*e) Foggy*

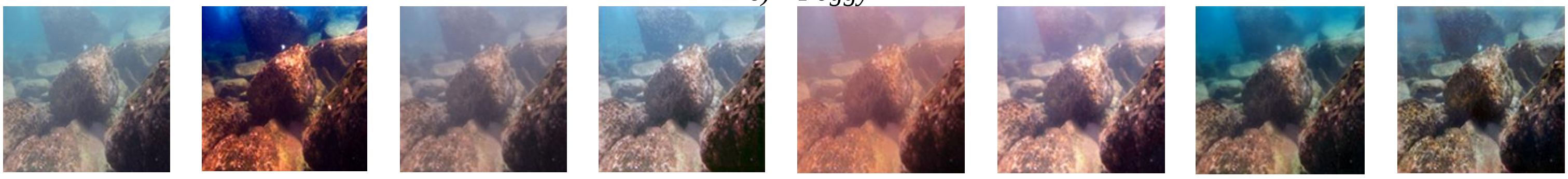

*f) Foggy*

**Figure 7.** Qualitative Comparison between Proposed UW-DiffPhys and Baseline UIE Methods on U45-Test

**Raw** | **UW-PhysCNN** | **UW-DDPM** | **UW-DiffPhys**

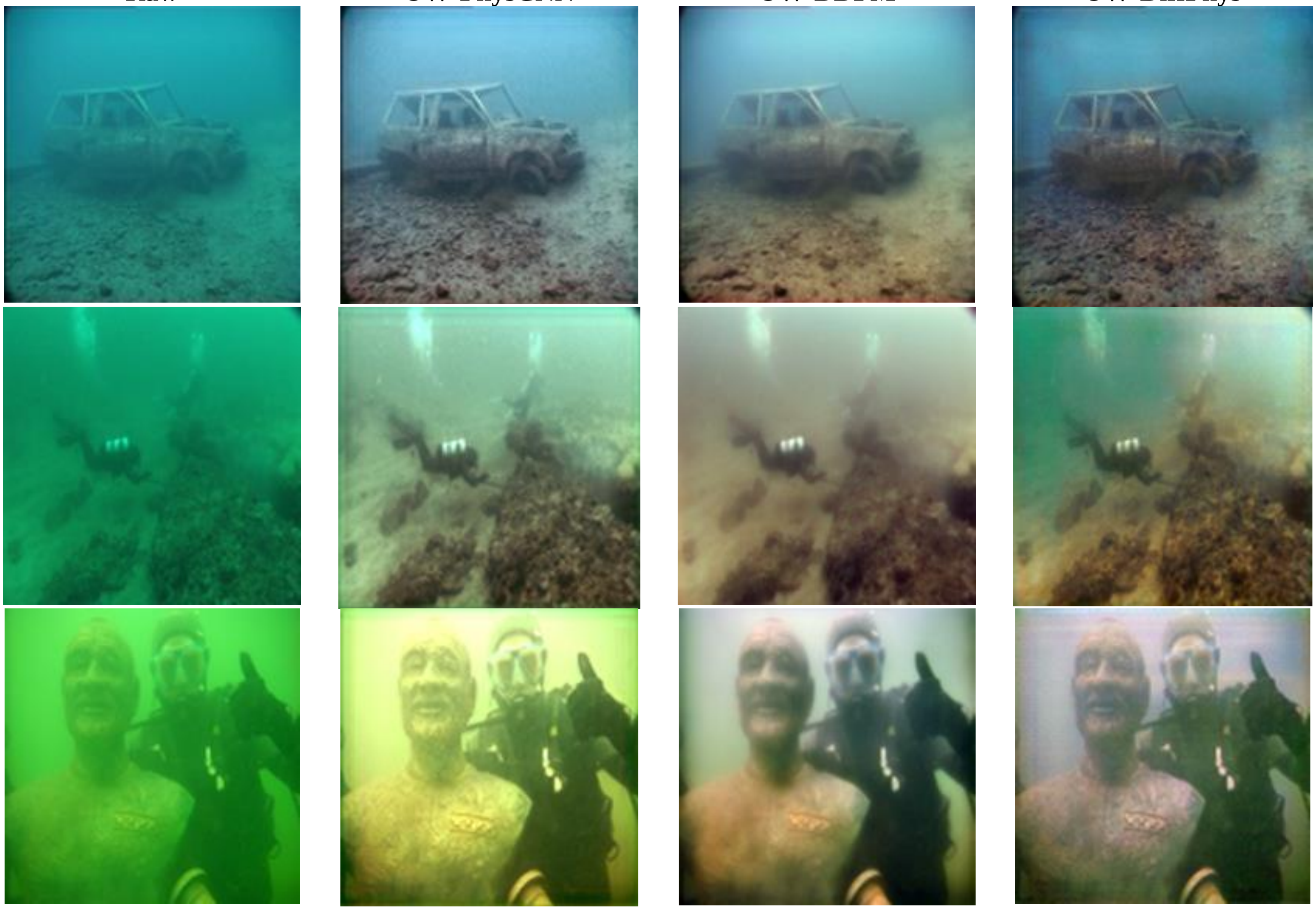

**Figure 8.** Enlarged examples of Proposed UW-DiffPhys, physical-based UIE (UW-PhysCNN), and diffusion-based UIE (UW-DDPM)

TABLE IV. QUALITATIVE COMPARISON BETWEEN THE PROPOSED UW-DIFFPHYS AND BASELINE UIE METHODS ON UIEB-TEST90, UIEB-TEST60, U45-TEST

| **Method** | **UIEB-Test90** | | | | **UIEB-Test60** | | **U45-Test** | |
|---|---|---|---|---|---|---|---|---|
| | **PSNR** | **SSIM** | **UCIQE** | **UIQM** | **UCIQE** | **UIQM** | **UCIQE** | **UIQM** |
| **Reference** | - | 1 | 0.6159 | 1.0876 | N/A | N/A | N/A | N/A |
| **Input** | 17.7723 | 0.7541 | 0.5448 | 0.6086 | 0.5253 | 0.1914 | 0.5328 | 0.5252 |
| **UDCP** | 13.5921 | 0.6413 | **0.618** | 0.6737 | 0.5682 | 0.1252 | <u>0.6005</u> | 0.4899 |
| **Shallow-UWNet** | 18.2951 | 0.7622 | 0.5437 | 0.5706 | 0.5208 | 0.1731 | 0.5299 | 0.4862 |
| **UW-PhysCNN** | <u>21.539</u> | <u>0.8644</u> | 0.5844 | **0.9502** | 0.5502 | **0.6533** | 0.5961 | **0.8823** |
| **MLFcGAN** | 16.6413 | 0.7365 | 0.576 | 0.5857 | 0.5405 | 0.1361 | 0.5519 | 0.4648 |
| **UWGAN** | 18.1082 | 0.8269 | 0.582 | 0.6638 | 0.5465 | 0.2707 | 0.569 | 0.5939 |
| **UW-DDPM** | **23.0567** | **0.8693** | <u>0.6041</u> | 0.8621 | **0.574** | 0.431 | **0.6029** | 0.8099 |
| **UW-DiffPhys** | 20.7419 | 0.8556 | 0.5922 | <u>0.9408</u> | <u>0.5691</u> | <u>0.6176</u> | 0.5863 | <u>0.8714</u> |

### 2. *Computational Complexity and Inference Time Comparison*

Table V shows the effectiveness of the proposed UW-DiffPhys to considerably reduce the computational complexity of UW-DDPM to approximately half, by replacing the U-Net for the Distribution Transformation task with light-computation underwater physical imaging components A-Net and T-Net, which are supported by the proposed Diffusion Underwater Physical Model. Additionally, the table also shows a significant decrease in inference time of UW-DiffPhys by utilizing a deterministic implicit sampling approach to require much fewer time steps (25 time steps), compared to UW-DDPM (1000 time steps).

TABLE V. COMPUTATIONAL COMPLEXITY AND INFERENCE TIME COMPARISON BETWEEN THE PROPOSED UW-DIFFPHYS AND UW-DDPM

| | **UW-DDPM** | **UW-DiffPhys** | | |
|---|---|---|---|---|
| | 2x U-Net | Denoising U-Net | A-Net T-Net | Total |
| **Computation Complexity** | 264.92 GFLOPs | 132.52 GFLOPs | 1598.06 MFLOPs | 134.12 GFLOPs |
| **Number of Parameters** | 171.21 M | 85.61 M | 46.14 k | 85.65 M |
| **Average Inference Time** | 19.55 s | 0.34 | | |

### 3. *Ablation Study*

As described in section III.A, the proposed UW-DiffPhys mainly contributes to the Distribution Transformation Network by integrating light-computation physical-based UIE networks (A-Net, T-Net) (1) into the existing high-computation Denoising Network $\theta$ (2). The integration requires finetuning $\theta$ to compensate for limited ability of A-Net, T-Net (3). This section shows the effectiveness of each component (1) (2), and investigates how (3) may affect the original quality performance of Denoising task.

Removing (1) from the contributions, the model is essentially UW-DDPM [19]. Qualitative results can be observed in Fig. 8 (third column), where the restoration of colors is desirable, but lacking restoration in scene details. Removing (2) from the contributions, the model is essentially UW-PhysCNN [33]. Fig. 8 (second column) reveals limited recovering of color, yet more scene details can be obtained. The final column in Fig. 8 indicates the proposed UW-DiffPhys can retain the two advantages from both previous models.

Regarding (3), fine-tuning the existing Denoising Network for the Distribution Transformation task, as shown in Eq. (25), may affect the overall quality performance during training. Figure 9 compares the loss performance of UW-DiffPhys (fine-tuning Denoising Network) and UW-DDPM (no fine-tuning Denoising Network) in terms of both the Distribution Transformation task and Denoising task. The vertical axis depicts loss value, whereas the horizontal axis depicts training iterations. In Fig. 9a, the green line is $Loss_{A,T}$, as shown in Eq. (26), the orange line is the proposed $Loss_{\phi}$ of UW-DiffPhys in Eq. (27), whereas the blue line is the original $Loss_{\phi}$ in UW-DDPM. In Fig. 9b, the purple line belongs to the proposed Denoising U-Net being fine-tuned for Distribution Transformation purpose, whereas the orange line is the original Denoising U-Net without being fine-tuned. Figure 9a shows that the proposed UW-DiffPhys maintains the same Distribution Transformation performance as UW-DDPM, while requiring only half of the computation complexity. However, the Denoising Network in UW-DiffPhys was fine-tuned to achieve the transformation task, and thus its original denoising performance was affected, as seen from an error gap (of value about 5e-4) in Fig. 9b. This drop in Denoising performance leads to a slight drop in quality measures such as PSNR, SSIM, and UCIQUE, as seen from Fig. 10a,b,c. Nevertheless, as seen from Fig. 10d, despite lower UIQM performance over earlier training iterations, the later iterations of UW-DiffPhys indicate that the proposal can outperform the existing solely diffusion-based UW-DDPM, by incorporating underwater physical imaging components to better remove the scattering effect and obtain enhanced images with higher sharpness and contrast.

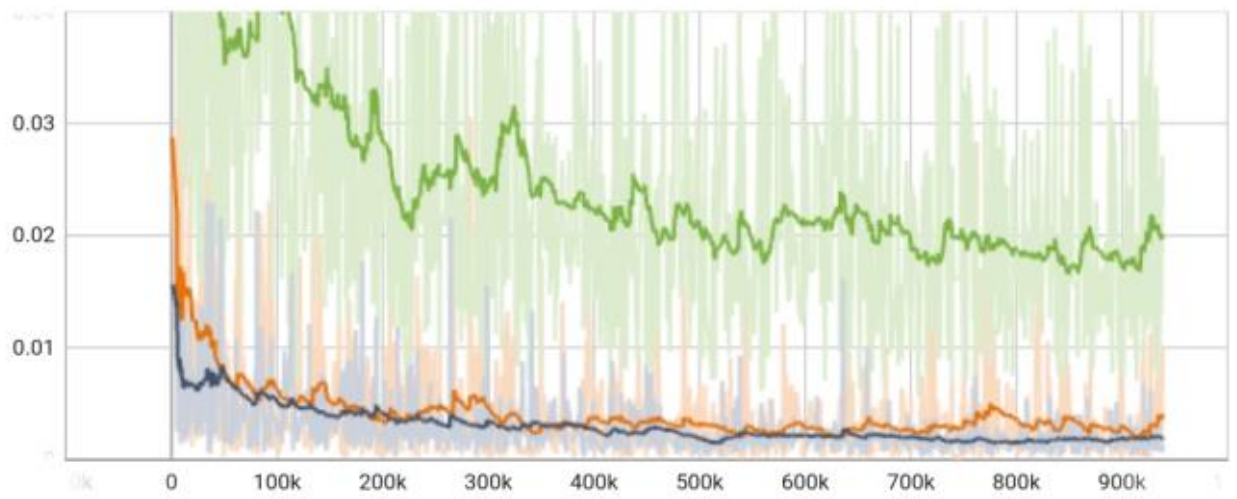

*a) Performance on Distribution Transformation Loss.*

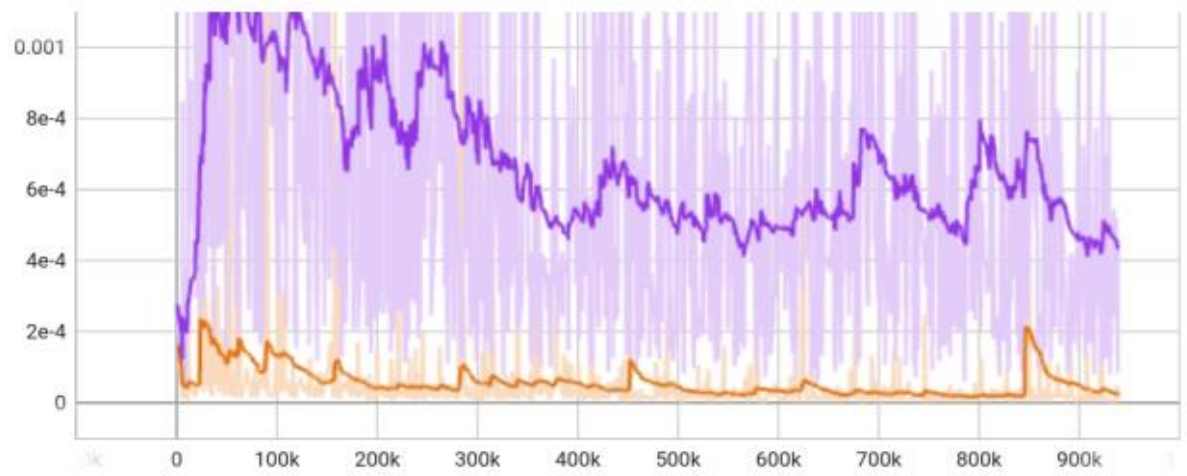

*b) Performance on Denoising Loss.*

**Figure 9.** Loss Performance between Proposed UW-DiffPhys and Original UW-DDPM [19].

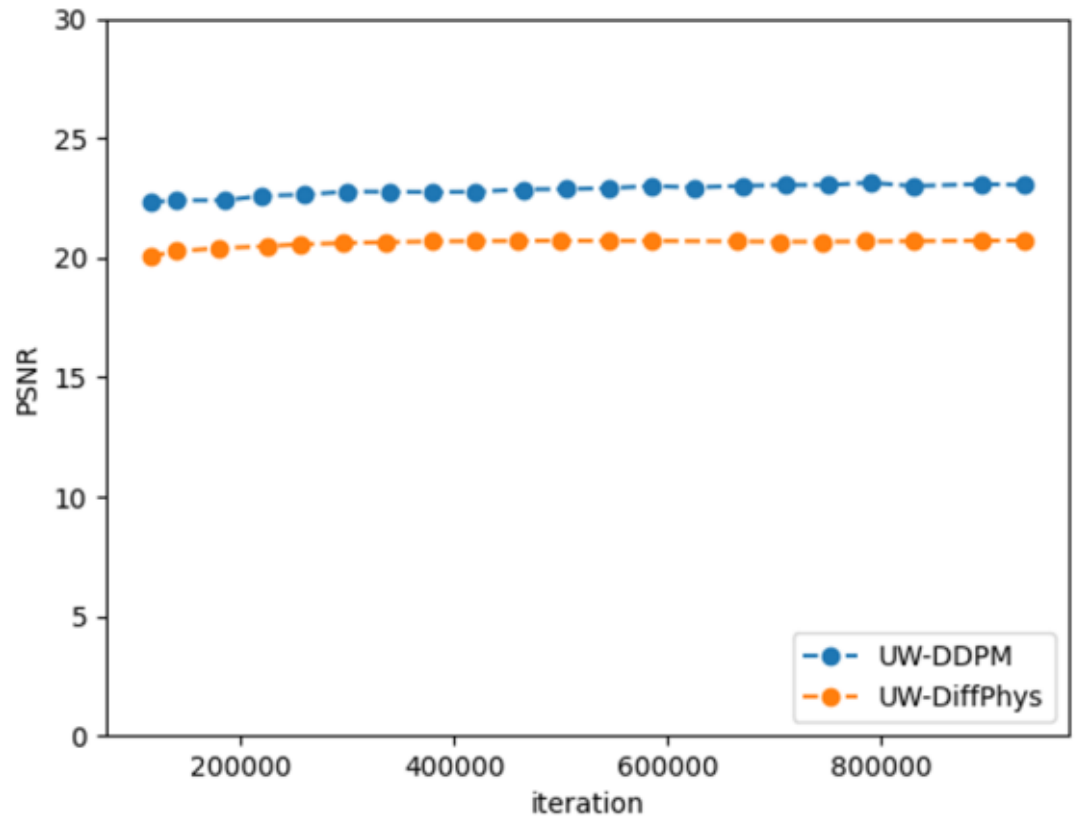

*a) PSNR*

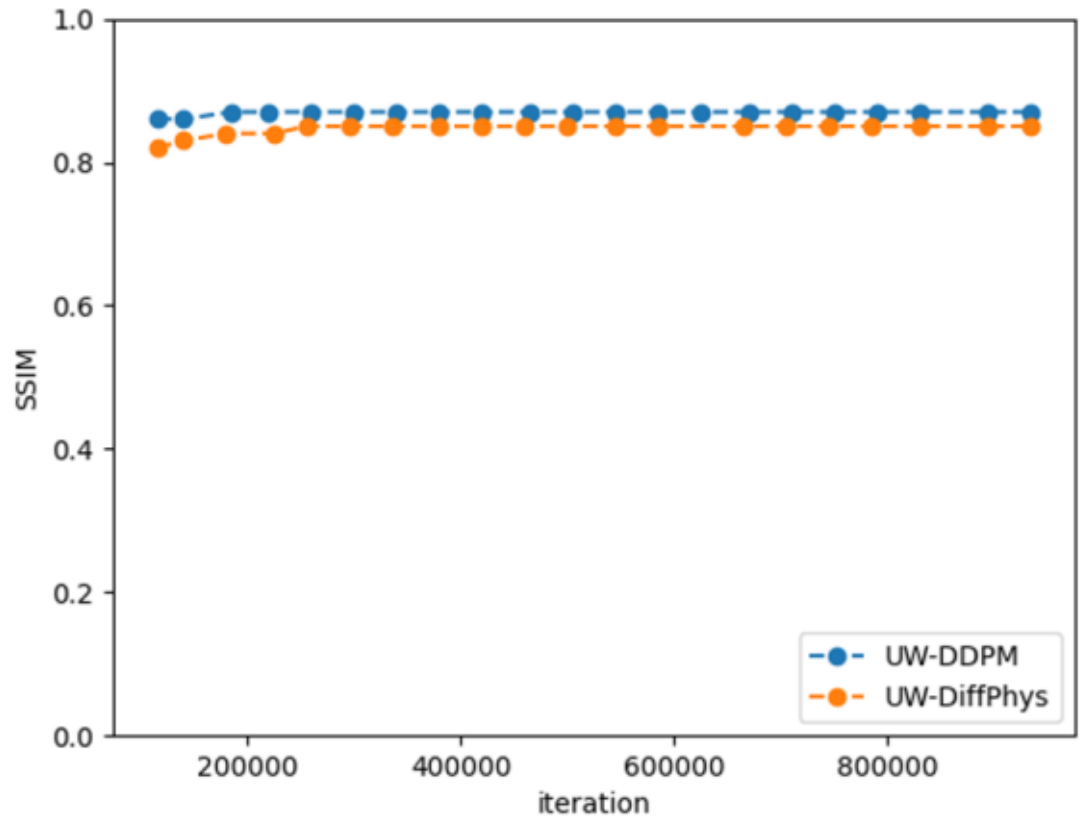

*b) SSIM*

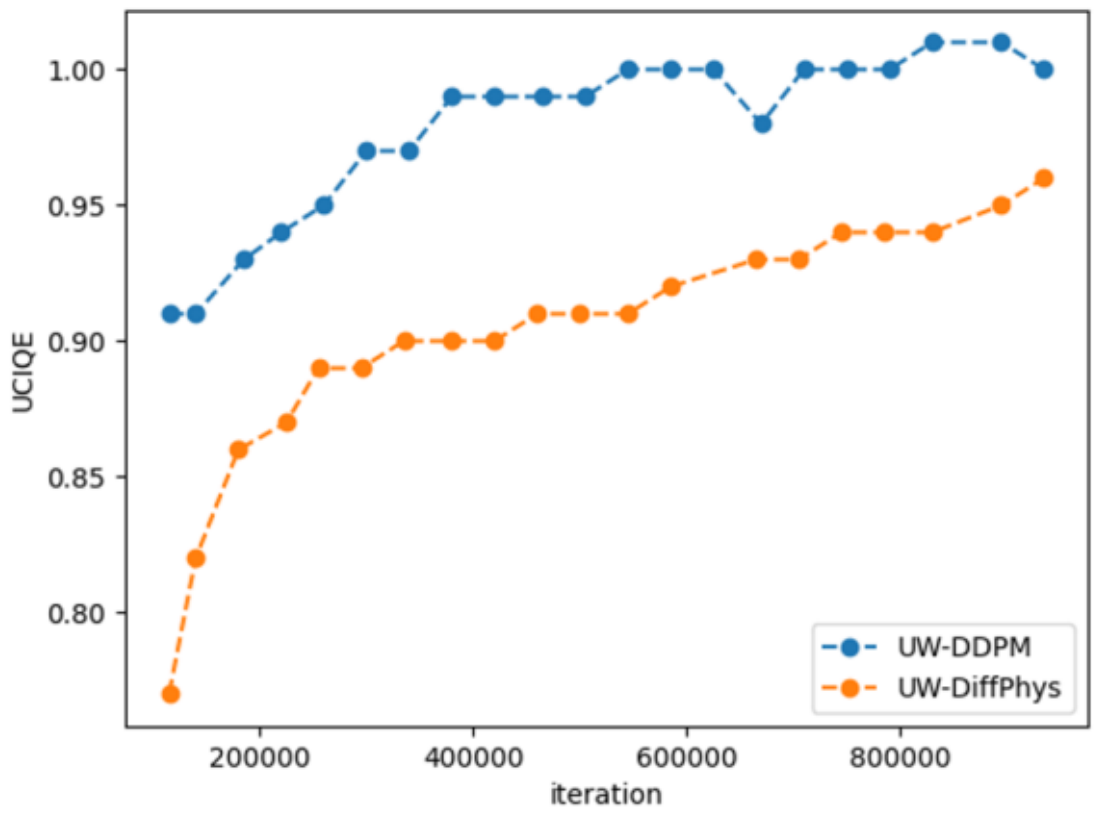

*c) UCIQUE*

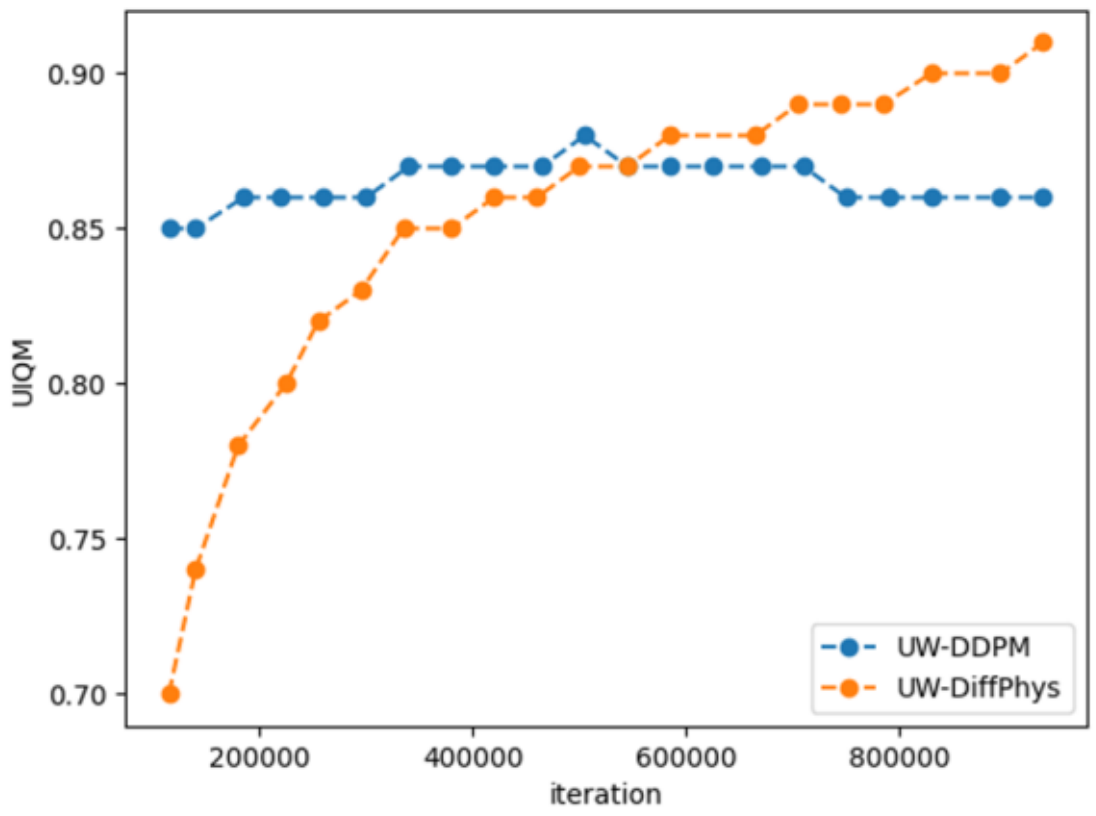

*d) UIQM*

**Figure 10.** Quality performance between Proposed UW-DiffPhys and Original UW-DDPM [19].

**Raw** **UW-DDPM** **UW-DiffPhys**

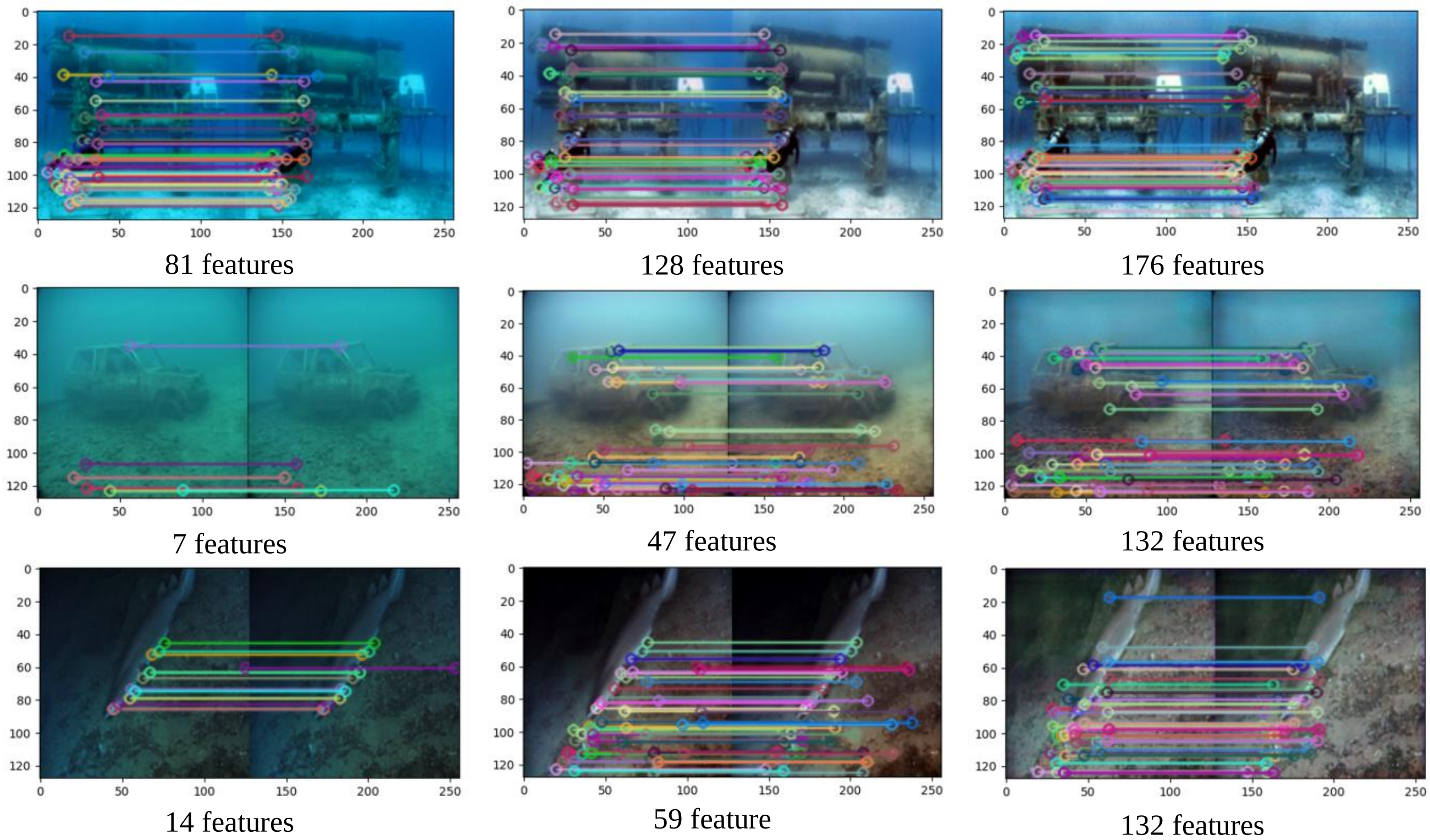

**Figure 11.** Feature points matching results on Raw underwater images, enhanced by UW-DDPM, and Proposed UW-DiffPhys

### 4. *Applicability on low-level AUV task*

This section provides example results when applying diffusion-based UIE methods for the low-level feature points matching task, which is an essential step for further high-level vision tasks like AUV localization and navigation. An enhanced image with high sharpness, and high contrast can effectively highlight edge and feature information, facilitating the features detection and matching process. Fig. 11 shows example results when applying SIFT feature matching method on three original degraded images, and those enhanced by diffusion-based UIE methods. The proposed UW-DiffPhys could consistently detect and match a higher number of feature points, compared to UW-DDPM and the Raw images.

## V. Discussion

This section discusses the effectiveness of the proposed UW-DiffPhys towards the research goal of this paper. The paper aims at improving real-time capability of the diffusion-based UIE on a low-cost AUV. The effectiveness of the proposal has been evaluated based on UIE metrics, computational complexity, inference time, and for a specific features matching task.

Regarding UIE metrics, the considerable increase of UW-DiffPhys in UIQM score when compared with UW-DDPM, and the outperformance in UCIQE score when compared with UW-PhysCNN, indicate that the proposed UW-DiffPhys can surpass the other diffusion-based UW-DDPM in the ability to bring more sharpness and contrast, while bringing fuller color enhancement than the physical-based UW-PhysCNN. On the decision between diffusion-based UW-DiffPhys and physical-based UW-PhysCNN, having a higher UCIQE makes UW-DiffPhys suitable for applications such as marine biology or environmental monitoring research [50]. Enhancing color accuracy and contrast may help in the identification and study of marine species or coral reefs health assessment, such as observing the color patterns in fishes or corals [50]. On the decision between the two diffusion-based UW-DiffPhys and UW-DDPM, having a higher overall quality UIQM score of sharpness, contrast, colorfulness makes UW-DiffPhys suitable for operation tasks of AUVs. The examples of low-level feature points matching task evaluated in section IV.4 indicate that UW-DiffPhys having higher sharpness and contrast could detect and match more features than UW-DDPM, despite UW-DDPM having higher colorfulness.

Regarding the computational complexity of the diffusion-based UIE, the proposed UW-DiffPhys has reduced the floating-point operations per second of UW-DDPM from 264.92 GFLOPs down to 134.12 GFLOPs, and the network parameters from 171.21 million to 85.65 million. Inspired by [62] [63], further investigation on other hardware constraints is needed to evaluate the deploy-ability of the proposed UW-DiffPhys. NVIDIA Jetson Nano has been widely utilized from mobile robots to low-cost AUV platforms for research and education purposes [51, 52]. The computational complexity limit on Jetson Nano is 472 GFLOPs [53]. Although both diffusion-based UIE methods can meet the computational requirement to run on the low-cost hardware, their existing performance is currently limited to small image size (128x128). For effective features matching task in visual simultaneous localization and mapping (SLAM) application of robots, a common image resolution at VGA level (640x480) [54] is necessary to obtain sufficient SLAM precision. However, the computational complexity (GFLOPs) is proportional to squared image resolution. This means that for a twice larger image size (256x256) but having not met the VGA level, both the diffusion-based UIE methods will increase the computational complexity by four times and exceed the requirement on low-cost Jetson Nano device (> 472 GFLOPs). Thus, further complexity reduction is needed to deploy the high-quality diffusion-based UIE models to process larger image resolutions on low-cost AUVs for essential operation tasks. Alternatively, a size-agnostic image enhancement approach [61] for diffusion-based UIE can be further investigated and adapted.

Regarding the inference time reduction, the original UW-DDPM operates at about 0.05 FPS (19.55s), meanwhile the proposed UW-DiffPhys can operate at approximately 3 FPS (0.34s). However, 3 FPS remains relatively slow for real-time applications, since real-time image analysis is generally considered to be at speed of 30 FPS or greater [55], whereas real-time localization of AUV should be larger than 15 FPS [56]. Nevertheless, the proposed UW-DiffPhys can still be applied for low frame rate applications such as underwater target tracking (1-5 FPS) [57, 58] or survey and mapping (5-10 FPS) [59] with further reduction of sampling steps. Since DDIM is among the first sampling techniques (commonly in 25 steps) to accelerate DDPM, more recent samplers such as UniPC (Unified Predictor-Corrector) [60] can be investigated and adapted in the future work to achieve high quality image enhancement in 5-10 steps.

## VI. Conclusion

This paper presents UW-DiffPhys, a novel approach to underwater image enhancement (UIE) that combines physical-based light-computation techniques with diffusion-based models to address the limitations of existing methods. By integrating light-computation physical-based UIE network components with a denoising U-Net, UW-DiffPhys reduces the computational complexity of the previous UW-DDPM framework while maintaining comparable performance. Additionally, the adoption of the Denoising Diffusion Implicit Model (DDIM) enables faster inference through non-Markovian sampling, considerably improving the real-time capability of the system. Experimental evaluations demonstrate that UW-DiffPhys achieves a notable reduction in computational complexity (approximately half) and inference time (less than 40 times the number of required time steps) compared to UW-DDPM. Although there is a slight decrease in some metrics such as PSNR, SSIM, and UCIQE, the proposed model shows a considerable increase in the overall underwater image quality UIQM performance, highlighting its effectiveness in enhancing underwater images. These contributions enable the high-quality diffusion-based image enhancement task to be applicable for low frame rate applications on low-cost AUVs processing limited image resolution. Future work is needed to continue improving real time capability for high frame rate applications of AUVs, and reducing computational complexity of both Denoising and Distribution Transformation networks of UW-DiffPhys to process larger image resolutions, for more accurate higher level AUV tasks.

## Conflict of Interest

The authors declare no conflict of interest.

## Author Contributions

Conceptualization, N.G.B.; Methodology, N.G.B.; Supervision, P.X.T. and E.K.; Writing—original draft, N.G.B, and C.M.T.; Writing—review and editing, P.X.T. and E.K. All authors have read and agreed to the published version of the manuscript.